\documentclass[10pt,twocolumn,letterpaper]{article}

\usepackage[pagenumbers]{cvpr} %

\usepackage{tabularx}

\renewcommand{\paragraph}[1]{\vspace{.5em}\noindent\textbf{#1.}}

\definecolor{cvprblue}{rgb}{0.21,0.49,0.74}
\usepackage[pagebackref,breaklinks,colorlinks,allcolors=cvprblue]{hyperref}

\def\confName{CVPR}
\def\confYear{2027}

\newcommand{\nOpenVLMs}{27}
\newcommand{\nAPIVLMs}{6}
\newcommand{\nAxes}{13}
\newcommand{\nAxesL}{11}
\newcommand{\nAxesS}{7}
\newcommand{\nBench}{19}
\newcommand{\nEval}{12,855}
\newcommand{\nSzero}{26}

\newcommand{\PairSelfPct}{52}
\newcommand{\PairPriorPct}{94}
\newcommand{\PairDirectPct}{42}
\newcommand{\nPairDirect}{25}
\newcommand{\PairAccCorr}{\ensuremath{+0.04}}

\newcommand{\PairsWorseMost}{16}
\newcommand{\PairsWorseTop}{10}
\newcommand{\FirstShownPct}{67}
\newcommand{\OverN}{23}
\newcommand{\LevelPredPct}{84}
\newcommand{\LevelActPct}{72}
\newcommand{\GeminiUnderPts}{12}
\newcommand{\SonnetUnderPts}{18}
\newcommand{\TerraPredPct}{86}
\newcommand{\TerraActPct}{87}
\newcommand{\TerraHiresActPct}{94}
\newcommand{\StatedOwn}{\ensuremath{+0.12}}
\newcommand{\StatedOwnLo}{\ensuremath{-0.03}}
\newcommand{\StatedOwnHi}{\ensuremath{+0.28}}
\newcommand{\StatedOwnPos}{14}
\newcommand{\AnswerOwn}{\ensuremath{+0.36}}

\newcommand{\AnswerCtl}{\ensuremath{-0.01}}

\newcommand{\StatedCtl}{\ensuremath{-0.01}}

\newcommand{\StatedDiff}{\ensuremath{+0.13}}
\newcommand{\StatedDiffLo}{\ensuremath{-0.002}}
\newcommand{\StatedDiffHi}{\ensuremath{+0.263}}

\newcommand{\GLMStatedOwn}{\ensuremath{+0.33}}
\newcommand{\GLMStatedCtl}{\ensuremath{-0.16}}

\newcommand{\StatedDiffThirteen}{\ensuremath{+0.12}}
\newcommand{\StatedDiffThirteenLo}{\ensuremath{+0.001}}
\newcommand{\StatedDiffThirteenHi}{\ensuremath{+0.236}}

\newcommand{\AnswerOwnFive}{\ensuremath{+0.26}}
\newcommand{\AnswerOwnFiveLo}{\ensuremath{+0.18}}
\newcommand{\AnswerOwnFiveHi}{\ensuremath{+0.34}}
\newcommand{\AnswerOwnFivePos}{23}
\newcommand{\LunaAnsFive}{\ensuremath{+0.46}}
\newcommand{\LunaAnsAll}{\ensuremath{+0.78}}
\newcommand{\LunaStatedL}{\ensuremath{-0.03}}
\newcommand{\TerraAnsFive}{\ensuremath{+0.16}}
\newcommand{\TerraAnsAll}{\ensuremath{+0.49}}
\newcommand{\TerraStatedL}{\ensuremath{-0.45}}
\newcommand{\AnsShortN}{6}
\newcommand{\AnsShortMean}{\ensuremath{+0.50}}
\newcommand{\AnsLenRemoved}{\ensuremath{+0.38}}
\newcommand{\AnsLenRemovedLo}{\ensuremath{+0.24}}
\newcommand{\AnsLenRemovedHi}{\ensuremath{+0.51}}
\newcommand{\PostOwn}{\ensuremath{+0.13}}

\newcommand{\nHsixb}{27}

\newcommand{\nHsixc}{25}

\newcommand{\TerraHsixcPts}{2}
\newcommand{\SonnetHsixcPts}{3}
\newcommand{\HsixcMainPts}{15}

\newcommand{\nObj}{27}
\newcommand{\BlurLargerDrop}{23}
\newcommand{\OcclLargerDrop}{18}
\newcommand{\nBlur}{25}
\newcommand{\ObjFlat}{22}
\newcommand{\BlurPredFall}{24}
\newcommand{\OcclPredFall}{23}

\newcommand{\DelegSelf}{0.02}
\newcommand{\DelegPrior}{0.84}

\newcommand{\FigOneSpatialPred}{80}
\newcommand{\FigOneSpatialAct}{88}
\newcommand{\FigOneVmathPred}{82}
\newcommand{\FigOneVmathAct}{52}
\newcommand{\FigOneSdModel}{0.074}
\newcommand{\FigOneSdMean}{0.043}

\newcommand{\RecErrNone}{22}
\newcommand{\RecErrFifteen}{12}
\newcommand{\RecImproveN}{25}

\newcommand{\RecNoMarks}{19}
\newcommand{\RecCount}{12}
\newcommand{\RecHigherPct}{89}

\newcommand{\RecSlopePos}{26}
\newcommand{\RecSlopeTen}{5.4}
\newcommand{\RecSlopeTenLo}{4.6}
\newcommand{\RecSlopeTenHi}{6.3}

\newcommand{\TrainRowsPerFold}{1{,}600}
\newcommand{\TrainPairLoPct}{48}
\newcommand{\TrainPairHiPct}{68}
\newcommand{\TrainPriorLoPct}{94}
\newcommand{\TrainPriorHiPct}{96}
\newcommand{\TrainOverKindsBeforeLo}{13}
\newcommand{\TrainOverKindsBeforeHi}{16}

\newcommand{\TrainOverKindsAfterHi}{2}
\newcommand{\TrainOverCondBeforeLo}{15}
\newcommand{\TrainOverCondBeforeHi}{33}
\newcommand{\TrainOverCondAfterLo}{3}
\newcommand{\TrainOverCondAfterHi}{9}
\newcommand{\TrainPairCondLo}{66}
\newcommand{\TrainPairCondHi}{70}
\newcommand{\TrainRevLo}{39}
\newcommand{\TrainRevHi}{63}
\newcommand{\TrainRefPairLo}{94}
\newcommand{\TrainRefPairHi}{98}

\newcommand{\TrainFarOverQwen}{7.5}

\newcommand{\TrainFarOverGemma}{9.1}

\newcommand{\TrainFarOverGLM}{7.6}
\newcommand{\TrainGLMPartOwnKinds}{\ensuremath{+0.36}}
\newcommand{\TrainGLMPartFarKinds}{\ensuremath{+0.40}}
\newcommand{\TrainGLMPartOwnCond}{\ensuremath{+0.47}}
\newcommand{\TrainGLMPartFarCond}{\ensuremath{+0.35}}
\newcommand{\DonorLikeGLM}{\ensuremath{-0.48}}

\newcommand{\HandoffBase}{0.10}
\newcommand{\HandoffOwn}{0.59}
\newcommand{\HandoffFar}{0.49}

\newcommand{\BudgetStepsSelf}{1.6}
\newcommand{\BudgetStepsOthers}{0.8}

\newcommand{\DropBase}{1.7}
\newcommand{\DropOwn}{1.0}
\newcommand{\DropFar}{1.2}
\newcommand{\DropRef}{0.7}
\newcommand{\ProbeKindsQwen}{\ensuremath{-}0.38}
\newcommand{\ProbeKindsGemma}{\ensuremath{-}0.24}
\newcommand{\ProbeKindsGLM}{\ensuremath{+0.22}}
\newcommand{\ProbeCondGLM}{\ensuremath{+0.44}}
\newcommand{\ProbeCondGLMQwenF}{\ensuremath{+0.35}}
\newcommand{\ProbeCondGLMGemmaF}{\ensuremath{+0.33}}

\newcommand{\nRes}{16}
\newcommand{\ResStatedOwn}{\ensuremath{+0.13}}
\newcommand{\ResOwn}{\ensuremath{+0.23}}
\newcommand{\ResOwnLo}{\ensuremath{+0.10}}
\newcommand{\ResOwnHi}{\ensuremath{+0.36}}
\newcommand{\ResCtl}{\ensuremath{-0.01}}
\newcommand{\ResStatedOwnLo}{\ensuremath{-0.07}}
\newcommand{\ResStatedOwnHi}{\ensuremath{+0.32}}
\newcommand{\ResBudgetGain}{\ensuremath{+0.01}}

\newcommand{\TOneVsReader}{\ensuremath{+0.04}}
\newcommand{\TOneVsCtl}{\ensuremath{+0.16}}
\newcommand{\DecAxis}{44}
\newcommand{\DecAxisLogit}{47}
\newcommand{\DecModel}{36}
\newcommand{\DecModelLogit}{36}
\newcommand{\DecRest}{20}
\newcommand{\DecRestLogit}{16}

\newcommand{\DecRestSd}{6.5}
\newcommand{\DecNoiseSd}{1.6}

\newcommand{\HaikuPairPct}{85}
\newcommand{\HaikuPriorPct}{70}
\newcommand{\HaikuDirectPct}{49}

\title{Do Vision-Language Models Know Their Own Visual Weaknesses?\\Probing Capability-Level Metacognition}

\author{
Dohoon Kim$^{1}$ \quad
    Taesup Moon$^{1,2}$\thanks{Corresponding author}
    \\
    $^{1}$ Department of Electrical and Computer Engineering, Seoul National University \\
    $^{2}$ ASRI/INMC/IPAI/AIIS, Seoul National University \\
    {\tt\small \{dohoon.kim, tsmoon\}@snu.ac.kr }
}

\begin{document}
\maketitle
\begin{abstract}
Metacognition includes knowing one's own strengths and weaknesses across kinds of tasks, not only judging whether a single answer is correct.
We test whether vision-language models (VLMs) know in advance which kinds of visual tasks and image conditions are hard for them, across a wide range of open and proprietary models.
Because some tasks are hard for every model, we separate what a model knows about this shared difficulty from what it knows about itself.
We find that VLMs know little in advance about their visual weaknesses.
They misjudge their overall accuracy and do not even know which kinds of tasks are hard for every model, and most misjudge how degraded images affect them.
We further train VLMs to predict their own accuracy, which teaches them their own level of accuracy on kinds unseen in training but not which kinds are hard for them.
However, information about these weaknesses appears once a model starts solving.
On a few unlabeled questions of a kind, the probability a model assigns to its own answers follows weaknesses particular to the model, across kinds of tasks and image conditions.
VLMs thus hold information about their own weaknesses that their stated predictions do not yet reflect.
By separating what is particular to a model from what holds for every model, our evaluation offers a test for methods that aim to teach VLMs their own weaknesses, and the answer probability offers them a signal to start from.
\end{abstract}

\section{Introduction}
\label{sec:intro}

Metacognition is the ability to recognize the limits of one's own abilities and a core component of intelligent behavior~\cite{flavell,nelson}.
Whether language models know what they know has become an active research question~\cite{kadavath,selfaware,binder,ackerman,survey}.
Most of this work asks whether a model can tell that its answer to a single question is correct~\cite{lin,tian,xiong,kapoor,han}.
However, the classic definition of metacognition also includes knowledge of one's own strengths and weaknesses across kinds of tasks~\cite{flavell}.
A vision-language model (VLM) may count objects reliably yet often misread small text in blurry images.
A VLM aware of these weaknesses could hand off document-reading tasks, or decide how sharp an image it needs, before trying a single question, and could recognize what it still needs to learn~\cite{selfimprove}.
This raises the central question of this paper.
Do VLMs know in advance which kinds of visual tasks and image conditions are hard for them?

\begin{figure*}[t]
  \centering
  \includegraphics[width=\linewidth]{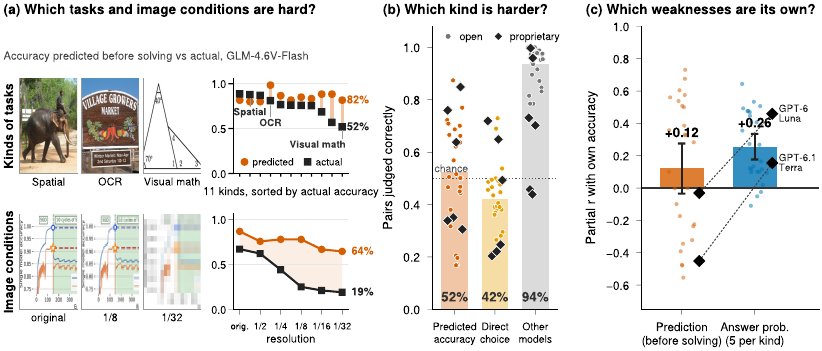}
  \caption{VLMs know little in advance about which visual tasks and image conditions are hard for them, yet the probability they assign to their own answers reveals weaknesses particular to the model.
  (a) Predicted and actual accuracy of GLM-4.6V-Flash, a typical open VLM (supplementary material), across \nAxesL{} kinds of tasks (top) and six resolutions of chart questions (bottom).
  It predicts similar accuracy for spatial relations and visual math (\FigOneSpatialPred\% and \FigOneVmathPred\%), whereas its actual accuracy is \FigOneSpatialAct\% and \FigOneVmathAct\%, and it underestimates the drop at lower resolution.
  (b) For pairs of kinds, how often a model's judgment matches the kind on which it is actually less accurate.
  Its predicted accuracies get this right for \PairSelfPct\% of pairs and its direct choices for \PairDirectPct\%, whereas the other models' average accuracy, computed for each model without it, gets \PairPriorPct\%, since what is hard for one model is mostly hard for every model.
  (c) How well the predictions and the answer probability show which kinds are harder for the model than for other models (partial r, \cref{sec:before-kinds}).
  Predictions before solving barely show this (\StatedOwn), whereas the answer probability on five unlabeled questions per kind does (\AnswerOwnFive).
  GPT-6 Luna and GPT-6.1 Terra show the same rise.
  Bars are means over open VLMs, dots single models, diamonds proprietary models, and error bars 95\% bootstrap intervals.}
  \label{fig:teaser}
\end{figure*}

Language models have been asked whether they will solve a given problem before attempting it, or whether to answer a query themselves or pass it to a stronger model~\cite{barkan,mirror,capsa,capcal}.
Yet these predictions concern single questions rather than kinds of tasks.
Visual tasks also add two factors that text tasks lack.
An image must first be perceived, and a model may not notice when it perceives it wrongly.
The image can also degrade, which makes the same question harder.
Evaluating such predictions is also not straightforward, because a model can know what is hard in two senses.
One is knowing which kinds of tasks are hard for every model, which is already useful, and the other is knowing which are hard for the model in particular.
A match between predicted and actual accuracy alone does not tell these apart.
So far, the two have been separated only for single questions in text models~\cite{moran,ashuach,binder}.
We therefore ask VLMs to predict their own accuracy on kinds of visual tasks and image conditions, and we separate how far their predictions follow what is hard for every model from how far they follow what is hard for the model itself.

We study these questions with \nOpenVLMs{} open VLMs and \nAPIVLMs{} proprietary models on \nAxes{} visual capability axes drawn from existing benchmarks.
Before a model answers any evaluation question, we ask how accurately it will answer each kind of task and how much its accuracy will drop as images degrade.
These predictions can go wrong in two ways.
They can be too high or too low overall, and they can misjudge which kinds of tasks are harder than others.
To judge which kinds a model finds hard, we use two standards.
The first is the other models' average accuracy, and predicting as well as this average shows that a model knows what is hard for every model.
The second is stricter.
A model shows knowledge of itself only if its predictions reveal something about the model that other models' predictions do not.

Our experiments lead to three findings.
First, VLMs know little in advance about their own visual weaknesses.
Their predictions are too high or too low overall, with most open models overestimating their accuracy and the strongest proprietary models underestimating it (\cref{fig:level}b).
They are also wrong about which kinds of tasks are hard for them, as for the model in \cref{fig:teaser}a (top).
Across models, their predictions pick the harder of two kinds in only \PairSelfPct\% of cases, whereas the other models' average accuracy picks it in \PairPriorPct\%, and no open model reaches that average (\cref{fig:teaser}b).
Handing off the kinds a model predicts to be hardest is therefore barely better than handing off at random.
Even the models that pick well rely on what is hard for every model, as their predictions barely follow where the model is stronger or weaker than other models (\cref{fig:teaser}c).
Most models likewise misjudge how degraded images and more objects reduce their accuracy (\cref{fig:teaser}a, bottom, and \cref{fig:image}).

Second, past outcomes tell a model its level of accuracy but do not teach it its weaknesses.
Given a record of its past answers marked correct or incorrect, a model mostly repeats how often it was right, and the same answers help little without the marks (\cref{fig:record}).
A model can also be trained to predict its accuracy, as prior work does for language models on single questions~\cite{kapoor,capsa,yax}.
We instead train VLMs to do so for kinds of tasks and image conditions, and test them on kinds unseen in training.
The trained models learn their own level of accuracy on these unseen kinds, while a model trained on another model's accuracy takes on that model's level instead.
Yet they pick the harder of two unseen kinds of tasks in only \TrainPairLoPct{} to \TrainPairHiPct\% of cases, whereas the other models' average accuracy picks it in \TrainPriorLoPct{} to \TrainPriorHiPct\%, and a model trained on another model's accuracy does just as well (\cref{tab:train}).
Training thus teaches a model its own level but not which kinds are hard for it.

Third, information about the model's own weaknesses appears once it starts solving (\cref{fig:teaser}c).
We let a model answer a few unlabeled questions of each kind and average the probability it assigns to its own answers~\cite{atc,fu}.
This answer probability follows the model's accuracy across kinds, including the weaknesses particular to the model, which other models' answer probabilities do not show.
This holds for \AnswerOwnFivePos{} of the \nSzero{} open models with five questions per kind and for both proprietary models that expose these probabilities.
It also follows the model's own accuracy loss at lower resolution.
Confidence that the model states in words right after each answer shows little of this.
VLMs thus hold information about their own weaknesses that they do not yet put into words.

Our contributions are summarized as follows:
\begin{itemize}
  \item We test whether VLMs know which kinds of visual tasks and image conditions are hard for them, across \nOpenVLMs{} open and \nAPIVLMs{} proprietary models and \nAxes{} capability axes, separating what is hard for every model from what is hard for the model itself. We release the question lists, every model's answers, and the scoring code.
  \item We find that VLMs, including frontier proprietary models, know little in advance about their visual weaknesses. Their predictions misjudge their overall accuracy and which kinds of tasks are hard, and most models misjudge how degraded images affect them.
  \item We show that past outcomes tell VLMs their level of accuracy but do not teach them their weaknesses. A model given a record of its answers mostly repeats how often it was right, and a model trained to predict its accuracy learns its own level on kinds unseen in training, but not which kinds are hard for it.
  \item We find that information about a model's own weaknesses is present in the probability it assigns to its own answers, which reveals weaknesses particular to the model across kinds of tasks and image conditions from a few unlabeled questions per kind.
\end{itemize}

\section{Related work}
\label{sec:related}
\paragraph{Metacognition in LLMs and VLMs}
Metacognition in language models has mostly been studied as the monitoring of single answers~\cite{survey}.
Models are asked for their confidence in an answer, in words or through token probabilities, and are evaluated or trained to be calibrated and to decline questions they would get wrong~\cite{kadavath,tian,xiong,lin,kapoor,rtuning,selfaware}.
Several studies measure this monitoring across domains, stages of answering, and internal states~\cite{ackerman,dmc,sdt,mmb,han,atlas,griot,automeco,jian,cash}.
Studies of VLMs likewise ask for confidence or abstention on single visual questions~\cite{whitehead,groot,vauq,recoverr,mmupd}, including whether a model knows what it can perceive in an image~\cite{mmsap} and how its confidence changes as visibility worsens~\cite{murky}.
Work that frames metacognition as knowing one's capabilities still asks for predictions on single items, such as whether the model will solve a given problem before attempting it or whether to answer or delegate a query~\cite{barkan,mirror,capsa,capcal}. 
LLMs can also name the skills a problem requires~\cite{didolkar}, which is knowledge about kinds of problems rather than about the model's accuracy on them.
Training models to predict their own accuracy has also been studied on single questions~\cite{steyvers,yax,esma}.
Other work finds where a model is weak without asking it, as benchmarks of the visual shortcomings of VLMs do~\cite{mmvp,blink,blind,vstar,cambrian}, or from its benchmark results or internal representations~\cite{evaltree,scan,gaps}.
We ask whether the model itself knows its accuracy on each kind of task and image condition, knowledge that the classic definition of metacognition includes~\cite{flavell}.

\paragraph{Shared difficulty and model-specific knowledge}
Some kinds of questions are hard for every model, so a prediction can match a model's accuracy without any knowledge of the model itself.
Recent work shows that such shared difficulty drives the confidence of language models~\cite{moran} and that much of what their hidden states reveal about their correctness is shared with other models~\cite{ashuach}.
Introspection tests ask whether a model predicts its own responses better than other models do~\cite{binder,song,singh}.
These studies separate shared from particular knowledge, but for single questions in text.
Internal signals such as token probabilities reveal errors that verbal confidence misses, in language models~\cite{azaria,orgad,steyversnmi} and in VLMs~\cite{lvlmkb,traps,ferdous}, and they estimate accuracy without labels~\cite{dengzheng,doc,atc,aotl,fu}, but without separating what they share with other models from what is particular to the model.
We draw this distinction at the level of kinds of tasks and image conditions, for both a model's stated predictions and its answer probabilities, against what the other models' accuracy and signals already reveal.

\section{Setup}
\label{sec:setup}
\begin{table}[t]
\caption{Capability axes. Items are evaluation questions. Scored and Pairs mark axes scored and compared in pairs.}
\label{tab:axes}
\centering
\scriptsize
\setlength{\tabcolsep}{2pt}
\begin{tabularx}{\columnwidth}{@{}>{\raggedright\arraybackslash}p{1.9cm}>{\raggedright\arraybackslash}Xrcc@{}}
\toprule
Axis & Source benchmarks & Items & Scored & Pairs \\
\midrule
  Attribute recognition & MMStar~\cite{mmstar}, MMVP~\cite{mmvp}, SEED-Bench~\cite{seedbench} & 862 & \checkmark & \checkmark \\
  Counting & BLINK~\cite{blink}, CV-Bench~\cite{cambrian}, TallyQA~\cite{tallyqa} & 1,340 & \checkmark & \checkmark \\
  Expert-level questions & MMMU-Pro~\cite{mmmupro}, MMStar~\cite{mmstar} & 926 & \checkmark & \checkmark \\
  High-resolution search & V*Bench~\cite{vstar}, HR-Bench~\cite{hrbench} & 418 & \checkmark & \checkmark \\
  Relative depth & BLINK~\cite{blink}, CV-Bench~\cite{cambrian} & 505 & \checkmark & \checkmark \\
  OCR & OCRBench~\cite{ocrbench} & 832 & \checkmark & \checkmark \\
  Visual math & MATH-Vision~\cite{mathvision}, MathVista~\cite{mathvista}, MathVerse~\cite{mathverse} & 827 & \checkmark & \checkmark \\
  3D distance & CV-Bench~\cite{cambrian} & 422 & \checkmark &  \\
  Chart reasoning & ChartQA~\cite{chartqa}, ChartQAPro~\cite{chartqapro} & 1,090 & \checkmark &  \\
  Chart value reading & ChartQAPro~\cite{chartqapro}, CharXiv~\cite{charxiv} & 2,310 & \checkmark &  \\
  Spatial relations & BLINK~\cite{blink}, CV-Bench~\cite{cambrian} & 546 & \checkmark &  \\
  Hallucination resistance & HallusionBench~\cite{hallusionbench}, AMBER~\cite{amber} & 2,064 &  &  \\
  Multi-view reasoning & MMSI-Bench~\cite{mmsibench} & 713 &  &  \\
\midrule
  Total & 19 benchmarks & 12,855 & 11 & 7 \\
\bottomrule
\end{tabularx}
\end{table}

We group visual questions into \nAxes{} capability axes drawn from \nBench{} existing benchmarks (\cref{tab:axes}).
Each axis is one kind of visual task and pools the questions of the benchmarks that test it.
We ask models about all axes but score their predictions only on the \nAxesL{} axes that rank the models reliably.
We evaluate \nOpenVLMs{} open VLMs from seven families, from 2B to 31B parameters, and \nAPIVLMs{} proprietary models (GPT-6 Luna, GPT-6.1 Terra, GPT-5-mini, Claude Haiku 5.5, Claude Sonnet 5.5, and Gemini 3.8 Flash), all on the same \nEval{} evaluation questions with the same prompts.
An analysis includes a model only if at least 90\% of its self-assessment replies follow the required answer format.
The full model list and the details of axis selection are in the supplementary material.

\section{Do VLMs know their weaknesses?}
\label{sec:before}
\subsection{Across kinds of visual tasks}
\label{sec:before-kinds}
\begin{figure}[t]
  \centering
  \includegraphics[width=\linewidth]{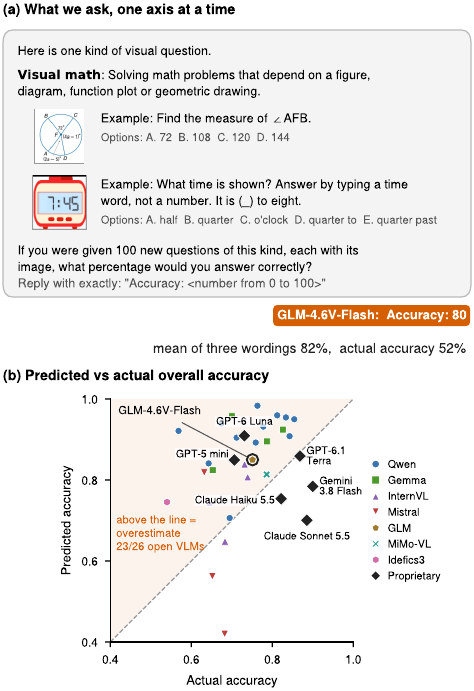}
  \caption{Predictions before solving.
  (a) The prompt for one axis, shown for visual math with its two example questions and the reply of GLM-4.6V-Flash.
  (b) Predicted against actual accuracy, both averaged over the \nAxesL{} scored axes.
  Each point is a model, colored by model family, with proprietary models as diamonds, and models above the dashed line overestimate their accuracy.}
  \label{fig:level}
\end{figure}

\paragraph{Setup}
Before a model answers any evaluation question, we ask it about one axis at a time (\cref{fig:level}a).
The prompt names the axis, defines it in one sentence, and shows two example questions with their images, taken from questions not used for evaluation.
It then asks what percentage of 100 new questions of this kind the model would answer correctly, and we average the answers over three wordings.
In a second format, we show two axes in the same way and ask on which the model would do better, six times per pair with both orders and the three wordings.
The predictions do not depend on whether the model itself answers the two examples correctly, and other example sets or a prompt that describes all \nAxes{} axes at once lead to the same conclusions (supplementary material).
We analyze the \nSzero{} open models that meet the format rule of \cref{sec:setup} and all \nAPIVLMs{} proprietary models.

\paragraph{Overall level}
We start with the overall level, which we measure as the mean of a model's predictions over the \nAxesL{} axes (\cref{fig:level}b).
Of the \nSzero{} open models, \OverN{} predict a higher accuracy than they reach, and on average they predict \LevelPredPct\% where they reach \LevelActPct\%.
The strongest proprietary models instead underestimate, with Gemini 3.8 Flash and Claude Sonnet 5.5 predicting \GeminiUnderPts{} and \SonnetUnderPts{} percentage points below their accuracy.
Getting the overall level right, however, does not mean knowing which kinds are hard.
GPT-6.1 Terra predicts \TerraPredPct\% on average, close to its actual \TerraActPct\%, yet the axis it expects to do worst on, high-resolution search, is one of its best, with \TerraHiresActPct\% accuracy.
Such errors cancel out in the average and show only when the axes are compared with one another.

\begin{figure}[t]
  \centering
  \includegraphics[width=0.9\linewidth]{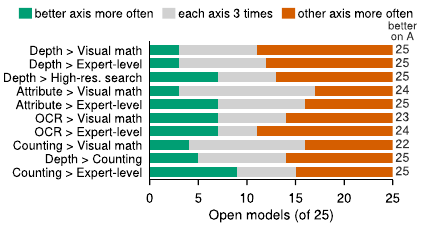}
  \caption{Direct choices between two axes for the 10 pairs with the largest gaps in accuracy.
  Each bar counts the \nPairDirect{} open models by whether they chose the axis they are better at more often, chose each axis three times, or chose the other axis more often.
  The number at the right is how many of them are more accurate on the first axis, and all 21 pairs are in the supplementary material.}
  \label{fig:pairs}
\end{figure}

\paragraph{Order of the axes}
We therefore compare the axes in pairs.
For each pair, the predictions count as correct if the axis with the lower predicted accuracy is the one on which the model is actually less accurate.
We use the 21 pairs of the \nAxesS{} axes on which a model's relative strength is consistent across benchmarks, each weighted by the gap in actual accuracy.
As a reference, we order the same axes by the mean accuracy of the open models other than the model itself.
This reference picks the harder axis in \PairPriorPct\% of cases, since what is hard for one model is mostly hard for every model.
The models' own predictions pick it in only \PairSelfPct\%, close to chance, and none of the \nSzero{} open models reaches the reference (\cref{fig:teaser}b).
More accurate models do not pick the harder axis more often (correlation \PairAccCorr{} across the open models).
Asking a model directly on which of two axes it would do better gives only \PairDirectPct\% correct choices, below chance.
In each of the \PairsWorseTop{} pairs with the largest gaps, more models choose the axis they are worse at than the axis they are better at (\cref{fig:pairs}).
Among the proprietary models, only Claude Haiku 5.5 beats the reference (\HaikuPairPct\% versus \HaikuPriorPct\%), but it chooses correctly in only \HaikuDirectPct\% when asked directly.

\begin{figure}[t]
  \centering
  \includegraphics[width=\linewidth]{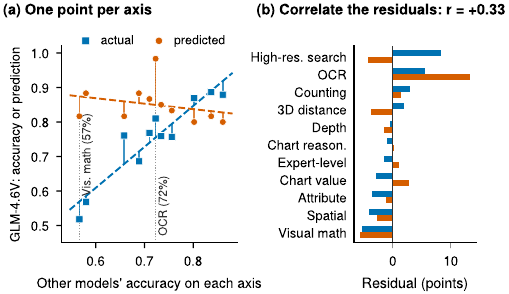}
  \caption{How we measure whether predictions follow where a model is stronger or weaker than other models, for GLM-4.6V-Flash.
  (a) Its accuracy and predictions on the \nAxesL{} scored axes against the other models' average accuracy, each with a fitted line whose vertical distances are the residuals.
  (b) The two residuals per axis, whose correlation (\GLMStatedOwn) is the partial correlation.}
  \label{fig:resid}
\end{figure}

\paragraph{Weaknesses particular to the model}
Picking the harder axis can rest on difficulty that every model shares, so we next ask whether the predictions reveal anything particular to the model.
Across the \nAxesL{} axes, we regress the model's accuracy on the other models' average accuracy, and the residuals show where the model is stronger or weaker than the other models suggest (\cref{fig:resid}a).
We regress its predictions in the same way and correlate the two sets of residuals (\cref{fig:resid}b).
This partial correlation is zero for predictions that follow only the shared difficulty or only the model's overall level.
It averages \StatedOwn{} [\StatedOwnLo, \StatedOwnHi] over the \nSzero{} open models, and \StatedOwnPos{} of them are above zero (\cref{fig:teaser}c).
As the second standard, we take the predictions that each other model made about itself, use them in place of the model's own, and compute the same partial correlation, averaged over the other models.
These predictions carry no knowledge of the model, so they show how high the score can be without it, and they score \StatedCtl{}.
The model's own predictions score \StatedDiff{} higher, but the 95\% interval of this difference, [\StatedDiffLo, \StatedDiffHi], includes zero.
Other ways of computing the partial correlation give the same conclusions (supplementary material).
\subsection{Under image conditions}
\label{sec:before-image}
\begin{figure}[t]
  \centering
  \includegraphics[width=0.875\linewidth]{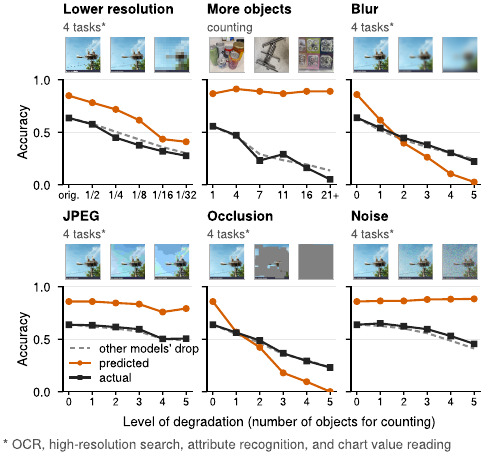}
  \caption{Predicted and actual accuracy of GLM-4.6V-Flash, the typical model of \cref{fig:teaser}, under image conditions.
  Images above each panel show the first, middle, and last level, and for counting they show 1--3, 11--15, and 21--50 objects.
  The dashed line is the reference, in which the model's accuracy falls from its first level as much as the other models' accuracy does on average.
  The mean over models is in the supplementary material.}
  \label{fig:image}
\end{figure}

\paragraph{Setup}
We next ask how a model expects its accuracy to change as images degrade.
The prompt shows three example questions of one axis with their images at one level of degradation and asks, as in \cref{sec:before-kinds}, what percentage of 100 such questions the model would answer correctly.
On text recognition, high-resolution search, attribute recognition, and chart value reading, we lower the resolution in six steps from the original to 1/32 of each side, and we also apply blur, JPEG compression, occlusion, and noise at five levels each.
On counting, we instead raise the number of objects in six ranges from 1--3 to more than 20.
At each level, we compare the drop the model predicts from the first level with its actual drop on the evaluation questions of that level, and as the reference we use the other open models' average actual drop.
We analyze all \nHsixb{} open models for resolution and object count, and \nHsixc{} for the other degradations because two models fail the inclusion rule of \cref{sec:setup} on them.

\paragraph{Results}
The predictions misjudge how much accuracy changes, too little under some conditions and too much under others (\cref{fig:image}).
As objects become more numerous, every open model predicts a smaller drop than it shows, and \ObjFlat{} of the \nObj{} models predict nearly the same accuracy for 21--50 objects as for 1--3, within 5 points.
Under blur and occlusion, nearly every model does expect its accuracy to fall (\BlurPredFall{} and \OcclPredFall{} of the \nBlur{}).
Yet most of them expect too large a drop, often down to zero (\BlurLargerDrop{} and \OcclLargerDrop{} models), whereas their accuracy only falls to the level of guessing on multiple-choice questions.
In every condition, the other models' average drop predicts the drop of nearly every model better than the model's own prediction does.
The proprietary models predict their drops better than the open models but still no better than the reference.
On the four degradations, the predicted drops of GPT-6.1 Terra and Claude Sonnet 5.5 are only \TerraHsixcPts{} and \SonnetHsixcPts{} points further from the actual drops than the reference is, against \HsixcMainPts{} points for the open models on average.
The exceptions are Claude Sonnet 5.5 and Gemini 3.8 Flash on counting, whose accuracy stays far above that of other models even with many objects.

\section{Can VLMs learn their weaknesses?}
\label{sec:learning}
VLMs thus know little in advance about their visual weaknesses.
We next ask whether their past outcomes can teach them.
We first give a model a record of its past answers in the prompt (\cref{sec:learning-context}) and then train it to predict its accuracy on each kind, with its past accuracy as the target (\cref{sec:learning-weights}).
\subsection{From a record in context}
\label{sec:learning-context}
\paragraph{Setup}
We repeat the prompt of \cref{sec:before-kinds} with a record of the model's past answers appended.
The record lists the text of 2, 5, or 15 questions of the same axis that the model answered earlier, each with the model's answer, whether it was correct, and the correct answer when it was wrong.
As a reference, we predict the model's accuracy on an axis by how often it was right in the record.
We analyze the \nSzero{} open models and measure the error as the mean absolute difference between predicted and actual accuracy over the \nAxesL{} axes.
Full prompts and the results for a record with images are in the supplementary material.

\begin{figure}[t]
  \centering
  \includegraphics[width=\linewidth]{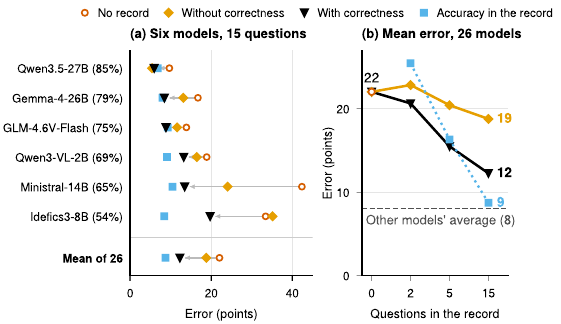}
  \caption{Predictions with a record of past answers.
  (a) Error of predicted accuracy with 15 questions per axis in the record, for six open models at every fifth rank of accuracy (in parentheses) and the mean of all \nSzero{}.
  Arrows go from no record to the record with correctness.
  (b) Mean error over the \nSzero{} models as the record grows, with the accuracy in the record (dotted) and the other models' average accuracy (dashed) used as predictions.}
  \label{fig:record}
\end{figure}
\paragraph{Results}
With 15 questions in the record, \RecImproveN{} of the \nSzero{} models predict more accurately than without a record (\cref{fig:record}a).
More accurate models become about as accurate as repeating the accuracy in their record, whereas less accurate models mostly predict above it.
On average, the error falls from \RecErrNone{} to \RecErrFifteen{} percentage points as the record grows to 15 questions and approaches that of repeating the accuracy in the record (dotted line in \cref{fig:record}b).
Without being told which answers were correct, the same answers reduce the error only to \RecNoMarks{} points.
The models thus gain from a record mainly because it tells them which of their answers were correct, and they can hardly tell this from the answers alone.
\subsection{From training on their outcomes}
\label{sec:learning-weights}
\paragraph{Setup}
A record shows a model how it did on one kind of task, whereas knowing one's weaknesses means anticipating them on kinds not yet tried.
We therefore ask whether training can teach a model this part of metacognition.
We train a model to predict its accuracy, with its past accuracy as the target, and test it only on kinds left out of training.
We fine-tune Qwen3.5-9B, Gemma-4-31B, and GLM-4.6V-Flash with LoRA~\cite{lora} on 31 kinds of tasks and the 109 image conditions of \cref{sec:before-image}.
The 31 kinds split the axes of \cref{tab:axes} by benchmark or benchmark subset, such as counting into BLINK, CV-Bench, and TallyQA.
We divide the kinds into four folds by axis and evaluate each kind only with the model trained without its axis.
This leaves 21 to 25 kinds of tasks for training in each fold.
Each training prompt is a prompt of \cref{sec:before-kinds,sec:before-image} with the model's accuracy as the answer, or asks on which of two kinds the model would do better.
We train one model per fold on about \TrainRowsPerFold{} such prompts for two epochs.
To separate what a model learns about itself from what any model's outcomes would teach, we also train each model in the same way on another model's accuracy~\cite{binder}.
Among the models within 10 points of its accuracy, we take the one whose strengths and weaknesses relative to the other models are least like those of the trained model.
We measure this likeness by the partial correlation of \cref{sec:before-kinds} between the two models' accuracies.
This is GLM-4.6V-Flash for Qwen3.5-9B and Gemma-4-31B, and Gemma-4-26B-A4B for GLM-4.6V-Flash.
Details are in the supplementary material.

\begin{table}[t]
\caption{Models trained to predict their accuracy, evaluated on kinds left out of training. The last row of each model is trained on another model's accuracy. Over. is the mean predicted minus actual accuracy in percentage points, best at 0. Pairs asks the model which of two unseen kinds it would do better on and gives the percentage of correct choices, weighted by the gap in accuracy. After the slash is the same score on only the pairs in which the model's actual order is the reverse of the order of the other models' average accuracy, where that average is always wrong. Part. is the partial correlation of \cref{sec:before-kinds}.}
\label{tab:train}
\centering
\footnotesize
\setlength{\tabcolsep}{3pt}
\resizebox{\linewidth}{!}{%
\begin{tabular}{@{}lrrrrrr@{}}
\toprule
 & \multicolumn{3}{c}{Kinds of tasks} & \multicolumn{3}{c}{Image conditions} \\
\cmidrule(lr){2-4}\cmidrule(l){5-7}
Trained on & Over.$\to$0 & Pairs$\uparrow$ & Part.$\uparrow$ & Over.$\to$0 & Pairs$\uparrow$ & Part.$\uparrow$ \\
\midrule
  \multicolumn{7}{@{}l}{\textit{Qwen3.5-9B}} \\
  \quad Untrained & +15.6 & 48\,/\,48 & +0.08 & +32.9 & 67\,/\,62 & +0.45 \\
  \quad Own & +2.3 & 53\,/\,42 & $-$0.49 & +7.3 & 68\,/\,61 & +0.49 \\
  \quad GLM-4.6V-Flash & $-$7.5 & 61\,/\,47 & $-$0.31 & +0.2 & 62\,/\,51 & +0.35 \\
\addlinespace[3pt]
  \multicolumn{7}{@{}l}{\textit{Gemma-4-31B}} \\
  \quad Untrained & +12.8 & 57\,/\,25 & +0.12 & +14.6 & 68\,/\,58 & +0.24 \\
  \quad Own & $-$0.9 & 48\,/\,39 & $-$0.11 & +8.9 & 70\,/\,64 & $-$0.02 \\
  \quad GLM-4.6V-Flash & $-$9.1 & 63\,/\,38 & $-$0.11 & $-$2.2 & 54\,/\,42 & $-$0.09 \\
\addlinespace[3pt]
  \multicolumn{7}{@{}l}{\textit{GLM-4.6V-Flash}} \\
  \quad Untrained & +14.8 & 50\,/\,57 & $-$0.01 & +18.9 & 59\,/\,61 & $-$0.02 \\
  \quad Own & +0.2 & 68\,/\,63 & +0.36 & +3.2 & 66\,/\,63 & +0.47 \\
  \quad Gemma-4-26B-A4B & +7.6 & 52\,/\,60 & +0.40 & +5.6 & 66\,/\,67 & +0.35 \\
\midrule
  Other models' average & & 94--96\,/\,0 & & & 98\,/\,0 & \\
\bottomrule
\end{tabular}}
\end{table}

\paragraph{Overall level}
Training on a model's own accuracy removes most of its overestimation on kinds left out of training (Over. in \cref{tab:train}).
On kinds of tasks, the mean prediction moves from \TrainOverKindsBeforeLo--\TrainOverKindsBeforeHi{} points above the actual accuracy to within \TrainOverKindsAfterHi{} points of it, and on image conditions from \TrainOverCondBeforeLo--\TrainOverCondBeforeHi{} to \TrainOverCondAfterLo--\TrainOverCondAfterHi{} points above it.
Training on another model's accuracy instead moves the predictions toward that model's level.
On kinds of tasks, Qwen3.5-9B and Gemma-4-31B trained on the accuracy of the less accurate GLM-4.6V-Flash underestimate their accuracy by \TrainFarOverQwen{} and \TrainFarOverGemma{} points, whereas GLM-4.6V-Flash trained on that of the more accurate Gemma-4-26B-A4B overestimates it by \TrainFarOverGLM{} points.
The level a model states thus comes from the outcomes it is trained on, and on kinds of tasks only its own outcomes give it its own level.

\paragraph{Which kinds are hard}
The trained models, however, still do not know which kinds are hard for them.
Asked which of two unseen kinds they would do better on, they choose correctly in only \TrainPairLoPct{} to \TrainPairHiPct\% of pairs of kinds of tasks and \TrainPairCondLo{} to \TrainPairCondHi\% of pairs of image conditions (Pairs).
The other models' average accuracy picks the right one in \TrainRefPairLo{} to \TrainRefPairHi\% of the same pairs, since most of the difference between kinds is shared by all models.
It is always wrong, however, on the pairs whose order is the reverse of that average (after the slash).
On these pairs of kinds of tasks, the trained models are right in only \TrainRevLo{} to \TrainRevHi\% of cases, around chance.
With training, Part. of GLM-4.6V-Flash rises from about 0 to \TrainGLMPartOwnKinds{} on kinds of tasks and \TrainGLMPartOwnCond{} on image conditions, while that of the other two models stays or falls (\cref{tab:train}).
Yet it rises about as much, to \TrainGLMPartFarKinds{} and \TrainGLMPartFarCond, when GLM-4.6V-Flash is trained on the accuracy of Gemma-4-26B-A4B, which tends to be strong where GLM-4.6V-Flash is weak and weak where it is strong (likeness \DonorLikeGLM).
Its own outcomes are therefore not what raises its Part.
For none of the three models does training on its own accuracy give a reliably larger Part. than training on another model's (95\% intervals; supplementary material).
Training on its own outcomes thus teaches a model its own level, but not which kinds are hard for it beyond what another model's outcomes teach.
\subsection{Decisions before and after training}
\label{sec:learning-decisions}
\begin{figure}[t]
  \centering
  \includegraphics[width=\linewidth]{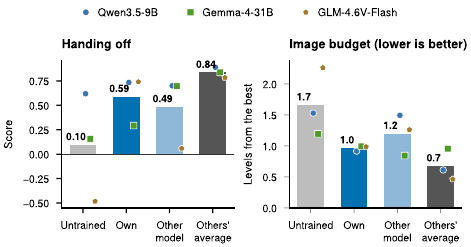}
  \caption{Decisions before and after training, on what each model was not trained on, with means over the three trained models as bars and single models as markers.
  Left, handing off the questions of the kinds predicted lowest, scored from $-1$ for the reverse of the actual order to 1 for the actual order.
  Right, choosing the cheapest image whose loss stays within a tolerated limit, measured in levels from the best choice.
  The other model is the one of \cref{sec:learning-weights}, or for the image budget the model whose drops are least like their own.}
  \label{fig:decisions}
\end{figure}
\paragraph{Setup}
We finally ask whether the predictions help a model make two decisions that need knowledge of its accuracy.
It can hand off the kinds it is worst at to a stronger system, and it can set an image budget by requesting a smaller image to save cost.
The two need different parts of this knowledge.
We evaluate each decision for the three trained models, before and after training, on kinds left out of training, and compare it with the same decision made from the other models' average accuracy.
Answering only the kinds above a threshold depends on the threshold chosen, and its results are in the supplementary material.

\paragraph{Handing off}
A model that can pass some questions to a stronger system should pass those of the kinds it is worst at, which requires knowing the order of the kinds.
We order the 11 axes by the predictions of \cref{sec:before-kinds} and hand off their 550 questions, 50 per axis, starting from the axis predicted lowest.
A handed-off question is taken as answered correctly, so the accuracy over all 550 questions rises as more of them are handed off.
We compute this accuracy for every number of questions handed off, from none to all, and average it.
We scale this average so that handing off in random order gives 0 on average, handing off in the order of the model's actual accuracy gives 1, and the reverse of that order gives $-1$.
Before training, the 26 open models score \DelegSelf{} on average, close to random order, whereas ordering by the other models' average accuracy scores \DelegPrior.
Training on their own accuracy raises the score of all three trained models, from \HandoffBase{} to \HandoffOwn{} on average, most for GLM-4.6V-Flash (\cref{fig:decisions}).
Training on another model's accuracy raises it to \HandoffFar{} on average, more than training on their own for Gemma-4-31B and less for GLM-4.6V-Flash, and both stay below the other models' average.
Training thus makes the model's predictions more useful for this decision, but they still serve it worse than the other models' average.

\paragraph{Image budget}
A model that can request a smaller image to save cost needs to know how much its accuracy drops at each size.
Resolution and JPEG compression are the two degradations that make an image cheaper.
On the four axes of \cref{sec:before-image}, we let a model choose the cheapest image whose predicted drop stays within the loss a user tolerates, for tolerated losses from 0 to 20 points.
We measure how many levels its choice lies from the cheapest image that actually stays within that loss.
Before training, the open models' choices lie \BudgetStepsSelf{} levels from the best on average, twice as far as choices made with the other models' average drops (\BudgetStepsOthers).
We also train the three models to predict their drop, in the same way as in \cref{sec:learning-weights}, which brings their choices from \DropBase{} to \DropOwn{} levels, most of the way to the other models' average (\DropRef; \cref{fig:decisions}).
Training on the drops of the model whose drops are least like their own brings them to \DropFar{} levels.

Training on past outcomes thus makes the predictions considerably more useful for both decisions, and training on the model's own outcomes helps slightly more on average than training on another model's.
Yet the results of other models still serve these decisions better.
To do better than these results, a model's own predictions would have to capture what is particular to it on top of the difficulty that all models share.

\section{Where is the model-specific information?}
\label{sec:where}
The stated and trained predictions examined so far reveal little about weaknesses particular to the model.
Neither does a linear probe on its hidden states before it solves (supplementary material).
We therefore look at what a model does while solving, the probability it assigns to its own answers.

\label{sec:where-confidence}
\paragraph{Setup}
For each axis, we let a model answer five questions without telling it whether it is right.
We take the mean probability it assigns to its own replies, with the probability of a reply taken as the geometric mean over its tokens.
Such scores are known to track accuracy on unlabeled questions~\cite{atc,fu}, and we ask whether they also track the weaknesses particular to the model.
We score them in the same way as the stated predictions, by the partial correlation over the 11 scored axes (\cref{sec:before-kinds}), with the other models' answer probabilities on the same questions as the second standard.
We analyze the \nSzero{} open models of \cref{sec:before-kinds} and the two proprietary models that return token probabilities, GPT-6 Luna and GPT-6.1 Terra.

\paragraph{Kinds of tasks}
The answer probability tracks where a model is stronger or weaker than other models, which its stated predictions miss (\cref{fig:teaser}c).
With five questions per axis, its partial correlation is \AnswerOwnFive{} [\AnswerOwnFiveLo, \AnswerOwnFiveHi] on average and above zero for \AnswerOwnFivePos{} of the \nSzero{} open models, against \StatedOwn{} for the stated predictions, and with all questions of an axis it reaches \AnswerOwn.
This does not depend on the length of the reply (supplementary material).
Another model's answer probabilities on the same questions do not follow this model's strengths and weaknesses (\AnswerCtl{} on average), so the signal is not a property of the questions themselves.
For the two proprietary models, the answer probability reaches \LunaAnsAll{} for GPT-6 Luna and \TerraAnsAll{} for GPT-6.1 Terra with all questions of an axis, and \LunaAnsFive{} and \TerraAnsFive{} with five, whereas their stated predictions give \LunaStatedL{} and \TerraStatedL.
A model thus carries, while solving, the information about its own weaknesses that its stated predictions lack, and five questions without known answers already reveal it for most models.
Asked right after answering how likely its answer is to be correct, it also reveals little of this information (\PostOwn), so current VLMs have this information but lack a way to put it into words.

\paragraph{Image conditions}
Among image conditions, we test lower resolution, a condition a model can choose itself to save cost (\cref{sec:learning-decisions}).
The answer probability also follows how much a model's accuracy drops as the resolution falls.
For the \nRes{} models whose number of image tokens falls with the resolution, its partial correlation with the model's own drop is \ResOwn{} [\ResOwnLo, \ResOwnHi] on average, against \ResCtl{} for other models' answer probabilities and \ResStatedOwn{} [\ResStatedOwnLo, \ResStatedOwnHi] for the stated drop.
The information that a model's stated predictions lack is thus present while it solves, across kinds of tasks and image resolutions alike.

\section{Conclusion}
\label{sec:discussion}
We asked whether VLMs know their own visual weaknesses, at the level of kinds of tasks and image conditions rather than single questions.
Before solving, they misjudge both their overall accuracy and which kinds of tasks and image conditions are hard for them.
Given a record of their past answers in context, their predictions move toward the accuracy in the record.
When we fine-tuned them to predict their accuracy on a kind, with their past accuracy as the target, they learned their own level of accuracy on kinds they were not trained on, but not which kinds are hard for them in particular.
Yet the probability a model assigns to its own answers on a few unlabeled questions follows these particular strengths and weaknesses.

By separating what is particular to a model from what holds for every model, our analysis shows which part of this knowledge current VLMs lack, which part records and training can supply, and where the missing part already lies.
This gives methods for teaching metacognition both a test and a signal to start from, and the next step is to teach models to report this signal and act on it.
A model that knows its own weaknesses could tell what it still needs to learn, so that we can train it there or it can improve itself.

\section*{LLM Usage Statement}
We used Claude (Anthropic) through Claude Code as an assistant for coding, experiments, and writing.
It wrote and ran code for the evaluations, analyses, and figures, suggested analyses, and drafted and edited the text of the paper, which the authors reviewed and revised.
The authors designed the study, made every analysis decision, and take full responsibility for the content.
The models evaluated in this paper are the subjects of the study and were not used to write it.

\clearpage
\maketitlesupplementary
\setcounter{section}{0}
\renewcommand{\thesection}{\Alph{section}}
The sections follow the main paper, with \cref{sec:suppl-fig1} for \cref{fig:teaser}, \cref{sec:suppl-axes} for \cref{sec:setup}, \cref{sec:suppl-before,sec:suppl-image} for \cref{sec:before}, \cref{sec:suppl-record,sec:suppl-train,sec:suppl-decisions} for \cref{sec:learning}, and \cref{sec:suppl-probe,sec:suppl-answer} for \cref{sec:where}; \cref{sec:suppl-think} reports models with reasoning turned on.

\section{Example model in Figure 1}
\label{sec:suppl-fig1}
\Cref{fig:teaser}(a) shows a single open VLM rather than the average over models, because averaging cancels the models' disagreements about which kinds are easy and makes the predictions look flatter than those of any one model: across the \nAxesL{} kinds, the standard deviation of a model's predictions has a median of \FigOneSdModel{}, against \FigOneSdMean{} for the averaged predictions.
We chose the model by a fixed rule, as the one closest to the median of the \nSzero{} open VLMs on four measures that the figure illustrates: how often its predictions order two kinds correctly, how much they vary across kinds, how much they overestimate accuracy, and how well they follow the drop at lower resolution.
GLM-4.6V-Flash is one of three models whose ranks are closest to the median in total, and the one among them whose predicted and actual drops at the lowest resolution are closest to the medians.

\section{Axes and models}
\label{sec:suppl-axes}
\subsection{Axes}
\begin{table*}[t]
\caption{The 13 capability axes. The definition is the sentence shown to the model in the prompt, Items is the number of evaluation questions, and Sep.\ is the standard deviation of accuracy across the 27 open models divided by the standard error of one model's accuracy (\cref{sec:suppl-axes}).}
\label{tab:suppl-axes}
\centering
\scriptsize
\setlength{\tabcolsep}{3pt}
\begin{tabularx}{\textwidth}{@{}>{\raggedright\arraybackslash}p{2.1cm}>{\raggedright\arraybackslash}X>{\raggedright\arraybackslash}p{5.4cm}rrl@{}}
\toprule
Axis & Definition & Source benchmarks (subsets) & Items & Sep. & Use \\
\midrule
  Attribute recognition & Recognizing the color, material, state or other attributes of objects. & MMStar~\cite{mmstar} (fine-grained perception); MMVP~\cite{mmvp}; SEED-Bench~\cite{seedbench} (instance attributes) & 862 & 2.7 & Scored, Pairs \\
  Counting & Counting how many objects of a given kind appear in an image. & BLINK~\cite{blink} (counting); CV-Bench~\cite{cambrian} (count); TallyQA~\cite{tallyqa} & 1,340 & 4.7 & Scored, Pairs \\
  Expert-level questions & Answering college-level questions in science, engineering, medicine or the humanities that use an image. & MMMU-Pro~\cite{mmmupro} (standard, four options); MMStar~\cite{mmstar} (reasoning) & 926 & 6.9 & Scored, Pairs \\
  High-resolution search & Locating a small object in a high-resolution image and answering about it. & V*Bench~\cite{vstar}; HR-Bench~\cite{hrbench} (4K); HR-Bench (8K) & 418 & 5.7 & Scored, Pairs \\
  OCR & Reading and transcribing the text in a given region of an image. & OCRBench v2~\cite{ocrbench} (recognition); OCRBench~\cite{ocrbench} (recognition) & 832 & 8.2 & Scored, Pairs \\
  Relative depth & Judging which of two marked points or objects is closer to the camera. & BLINK~\cite{blink} (relative depth); CV-Bench~\cite{cambrian} (depth) & 505 & 5.4 & Scored, Pairs \\
  Visual math & Solving math problems that depend on a figure, diagram, function plot or geometric drawing. & MATH-Vision~\cite{mathvision} (test); MathVista~\cite{mathvista} (multiple choice); MathVerse~\cite{mathverse} (vision-dominant) & 827 & 8.9 & Scored, Pairs \\
  3D distance & Judging which of two objects is closer to a third object in real 3D space. & CV-Bench~\cite{cambrian} (distance) & 422 & 4.6 & Scored \\
  Chart reasoning & Comparing values, judging trends or checking a claim using a chart. & ChartQA~\cite{chartqa} (human-written); ChartQAPro~\cite{chartqapro} (fact-checking); ChartQAPro (multiple choice) & 1,090 & 6.5 & Scored \\
  Chart value reading & Reading a specific value, label or title off a chart. & ChartQAPro~\cite{chartqapro} (factoid); CharXiv~\cite{charxiv} (descriptive, numeric); CharXiv (descriptive, text); CharXiv (descriptive, yes/no) & 2,310 & 14.0 & Scored \\
  Spatial relations & Judging where objects are relative to each other in an image (left of, above, and so on). & BLINK~\cite{blink} (spatial relation); CV-Bench~\cite{cambrian} (relation) & 546 & 5.0 & Scored \\
  Hallucination resistance & Not claiming things that are not in the image, including when the image is edited or misleading. & HallusionBench~\cite{hallusionbench} (image questions); AMBER~\cite{amber} & 2,064 & 5.0 & None \\
  Multi-view reasoning & Relating object positions and camera motion across several photos of the same scene. & MMSI-Bench~\cite{mmsibench} & 713 & 1.5 & None \\
\bottomrule
\end{tabularx}
\end{table*}

We keep an axis only if its sources rank the \nOpenVLMs{} open models alike (\cref{tab:suppl-axes}).
For each pair of sources, we correlate the models' accuracies on the two sources and correct the correlation for the noise of the finite number of questions, and we drop the axis only when the whole 95\% bootstrap interval of this corrected correlation lies below 0.7.
Because the strongest models win on every benchmark, two sources can agree through overall accuracy alone, so we repeat the check after subtracting each model's mean accuracy over all sources.
An axis must also separate the models, with a standard deviation of accuracy across models at least twice the standard error of one model's accuracy from the finite number of questions.
Of the \nAxes{} axes, \nAxesL{} are scored, and the \nAxesS{} whose sources still agree after the subtraction are also compared in pairs.
3D distance has a single source, and the sources of chart reasoning, chart value reading, and spatial relations no longer agree after the subtraction, so these four are scored but not compared in pairs.
The models are asked about hallucination resistance and multi-view reasoning as well, but these two axes are not scored, since the sources of the first disagree and the second does not separate the models.

\subsection{Models and decoding}
\begin{table}[t]
\caption{The 27 open VLMs by family, with their Hugging Face ids, and the 6 proprietary models, with their API names.}
\label{tab:suppl-models}
\centering
\scriptsize
\setlength{\tabcolsep}{4pt}
\begin{tabular}{@{}ll@{}}
\toprule
Family & Model \\
\midrule
  GLM & zai-org/GLM-4.1V-9B-Thinking \\
   & zai-org/GLM-4.6V-Flash \\
  Gemma & google/gemma-4-E2B-it \\
   & google/gemma-4-E4B-it \\
   & google/gemma-4-26B-A4B-it \\
   & google/gemma-4-31B-it \\
  Idefics & HuggingFaceM4/Idefics3-8B-Llama3 \\
  InternVL & OpenGVLab/InternVL3-2B-hf \\
   & OpenGVLab/InternVL3\_5-2B-HF \\
   & OpenGVLab/InternVL3-8B-hf \\
   & OpenGVLab/InternVL3\_5-8B-hf \\
  MiMo & XiaomiMiMo/MiMo-VL-7B-RL-2508 \\
  Mistral & mistralai/Ministral-3-8B-Instruct-2512-BF16 \\
   & mistralai/Ministral-3-14B-Instruct-2512-BF16 \\
   & mistralai/Mistral-Small-3.1-24B-Instruct-2503 \\
  Qwen & Qwen/Qwen2-VL-2B-Instruct \\
   & Qwen/Qwen3-VL-2B-Instruct \\
   & Qwen/Qwen3.5-2B \\
   & Qwen/Qwen2.5-VL-3B-Instruct \\
   & Qwen/Qwen3-VL-4B-Instruct \\
   & Qwen/Qwen3.5-4B \\
   & Qwen/Qwen2-VL-7B-Instruct \\
   & Qwen/Qwen2.5-VL-7B-Instruct \\
   & Qwen/Qwen3-VL-8B-Instruct \\
   & Qwen/Qwen3.5-9B \\
   & Qwen/Qwen3.5-27B \\
   & Qwen/Qwen3.8-27B \\
  \midrule
  OpenAI & gpt-6-luna \\
   & gpt-6.1-terra \\
   & gpt-5-mini-2025-08-07 \\
  Anthropic & anthropic/claude-haiku-5.5 \\
   & anthropic/claude-sonnet-5.5 \\
  Google & google/gemini-3.8-flash \\
\bottomrule
\end{tabular}
\end{table}

\Cref{tab:suppl-models} lists the models.
The open models run in Hugging Face Transformers with greedy decoding, with reasoning turned off where the chat template has a switch, and with at most 2{,}048 new tokens, or 4{,}096 for models that reason in their replies.
When a reply ends without an answer, we add ``Answer:'' to it and let the model continue for a few tokens.
The trained models of \cref{sec:learning-weights,sec:learning-decisions} and their untrained versions run in vLLM, also with greedy decoding.
\begin{table}[t]
\caption{Settings of the proprietary models. Reasoning is turned off where the API allows it and otherwise set to its lowest level. Temp.\ is the sampling temperature, Tokens the maximum number of new tokens, and Image the size limit in megapixels, above which images are scaled down to just below it.}
\label{tab:suppl-api}
\centering
\scriptsize
\setlength{\tabcolsep}{2pt}
\begin{tabular}{@{}llllrr@{}}
\toprule
Model & Access & Reasoning & Temp. & Tokens & Image \\
\midrule
  GPT-6 Luna & OpenAI & off & 0 & 4,096 & 16.9 \\
  GPT-6.1 Terra & OpenAI & off & 0 & 4,096 & 16.9 \\
  GPT-5-mini & OpenAI & minimal & default & 4,096 & 16.9 \\
  Claude Haiku 5.5 & OpenRouter & off & default & 4,096 & 1.2 \\
  Claude Sonnet 5.5 & OpenRouter & minimal & default & 8,192 & 3.7 \\
  Gemini 3.8 Flash & OpenRouter & minimal & 0 & 8,192 & 16.9 \\
\bottomrule
\end{tabular}
\end{table}

The proprietary models run through the OpenAI API or OpenRouter with the settings of \cref{tab:suppl-api}.

An analysis includes a model only if at least 90\% of its replies follow the required answer format.
This rule leaves out GLM-4.1V-9B-Thinking from the predictions of \cref{sec:before-kinds,sec:learning-context}, Qwen2-VL-2B as well from the direct choices between two axes, and both Qwen2-VL models from the four degradations of \cref{sec:before-image}.

\subsection{Actual profiles and shared difficulty}
\begin{figure*}[t]
  \centering
  \includegraphics[width=\textwidth]{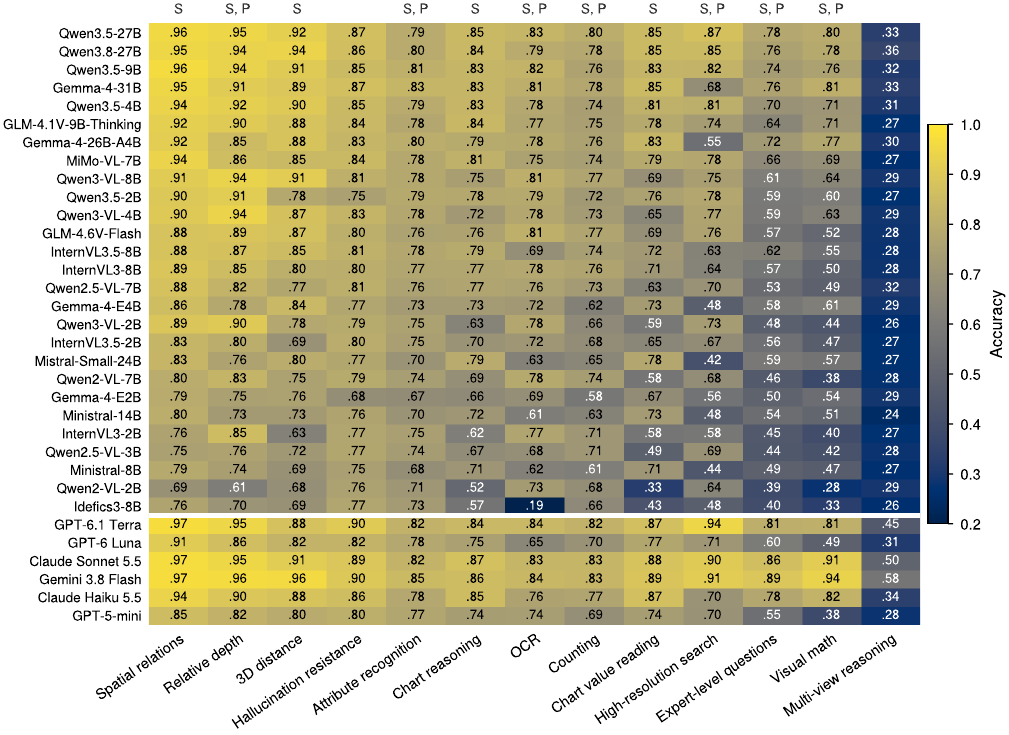}
  \caption{Accuracy of every model on every axis, on the evaluation questions.
  Open models are ordered by mean accuracy over the \nAxesL{} scored axes and shown above the proprietary models, and axes are ordered from easiest to hardest for the open models.
  S marks the scored axes and P those also compared in pairs.}
  \label{fig:suppl-profiles}
\end{figure*}
\Cref{fig:suppl-profiles} shows the accuracy of every model on every axis.
We split the variation in the open models' accuracies on the \nAxesL{} scored axes into three parts, as in a two-way analysis of variance.
The first is the difficulty of the axis, given by the mean accuracy of all models on it, and the second is the overall level of the model, given by its mean accuracy over all axes.
The third is the remainder, which shows where a model does better or worse on an axis than these two means suggest.
The difficulty of the axis accounts for \DecAxis\% of the variation, the level of the model for \DecModel\%, and the remainder for \DecRest\% (\DecAxisLogit\%, \DecModelLogit\%, and \DecRestLogit\% on the logit scale).
The remainder has a standard deviation of \DecRestSd{} points.
Because each accuracy is measured on a finite number of questions, this would be about \DecNoiseSd{} points even if all models had the same strengths and weaknesses, so the remainder mostly reflects real differences between models.
What is hard for one model is thus mostly hard for every model, while a smaller but reliable part is particular to each model.

\section{Predictions across kinds of tasks}
\label{sec:suppl-before}
\subsection{Prompts and examples}
\label{sec:suppl-prompts}
\begin{figure}[t]
  \centering\scriptsize
  \fbox{\begin{minipage}{0.95\linewidth}\ttfamily\raggedright
  Here is one kind of visual question.\\[3pt]
  Counting objects: Counting how many objects of a given kind appear in an image.\\
  \quad Example: How many skiers are wearing red?\ \ Options: A. 2\ \ B. 3\ \ C. 0\ \ D. 1\\
  \quad Image: [image]\\
  \quad Example: How many toilets are in the image?\ \ Options: A. 3\ \ B. 0\ \ C. 1\ \ D. 2\\
  \quad Image: [image]\\[3pt]
  If you were given 100 new questions of this kind, each with its image, what percentage would you answer correctly?\\
  Reply with exactly: "Accuracy: <number from 0 to 100>"
  \end{minipage}}\\[2pt]
  (a) Single-axis prompt, first wording.\\[5pt]
  \fbox{\begin{minipage}{0.95\linewidth}\ttfamily\raggedright
  Consider two kinds of visual questions.\\[3pt]
  X. Counting objects: Counting how many objects of a given kind appear in an image.\\
  \quad Example: \dots\ \ Image: [image]\ \ (two examples, as in (a))\\
  Y. Visual math: Solving math problems that depend on a figure, diagram, function plot or geometric drawing.\\
  \quad Example: \dots\ \ Image: [image]\\[3pt]
  On which kind would you answer a larger share of new questions correctly?\\
  Reply with exactly "Answer: X" or "Answer: Y".
  \end{minipage}}\\[2pt]
  (b) Direct choice between two axes, first wording.
  \caption{The prompts of \cref{sec:before-kinds}. [image] marks where the image is sent, at its original size.
  The second wording opens with ``Think about how well you handle the following type of image-based question.'' and asks ``What share of new questions of this type would you get right?'', and the third opens with ``You will be tested on the following category of visual questions.'' and asks ``Predict your accuracy on this category.''
  For direct choices, the second wording opens with ``Here are two types of image-based questions.'' and asks ``Which type are you better at, meaning you would get a higher percentage of new questions of that type right?'', and the third opens with ``You will be tested on two categories of visual questions.'' and asks ``On which category will your accuracy be higher?''}
  \label{fig:suppl-prompts}
\end{figure}
\begin{figure*}[t]
  \centering
  \includegraphics[width=\textwidth]{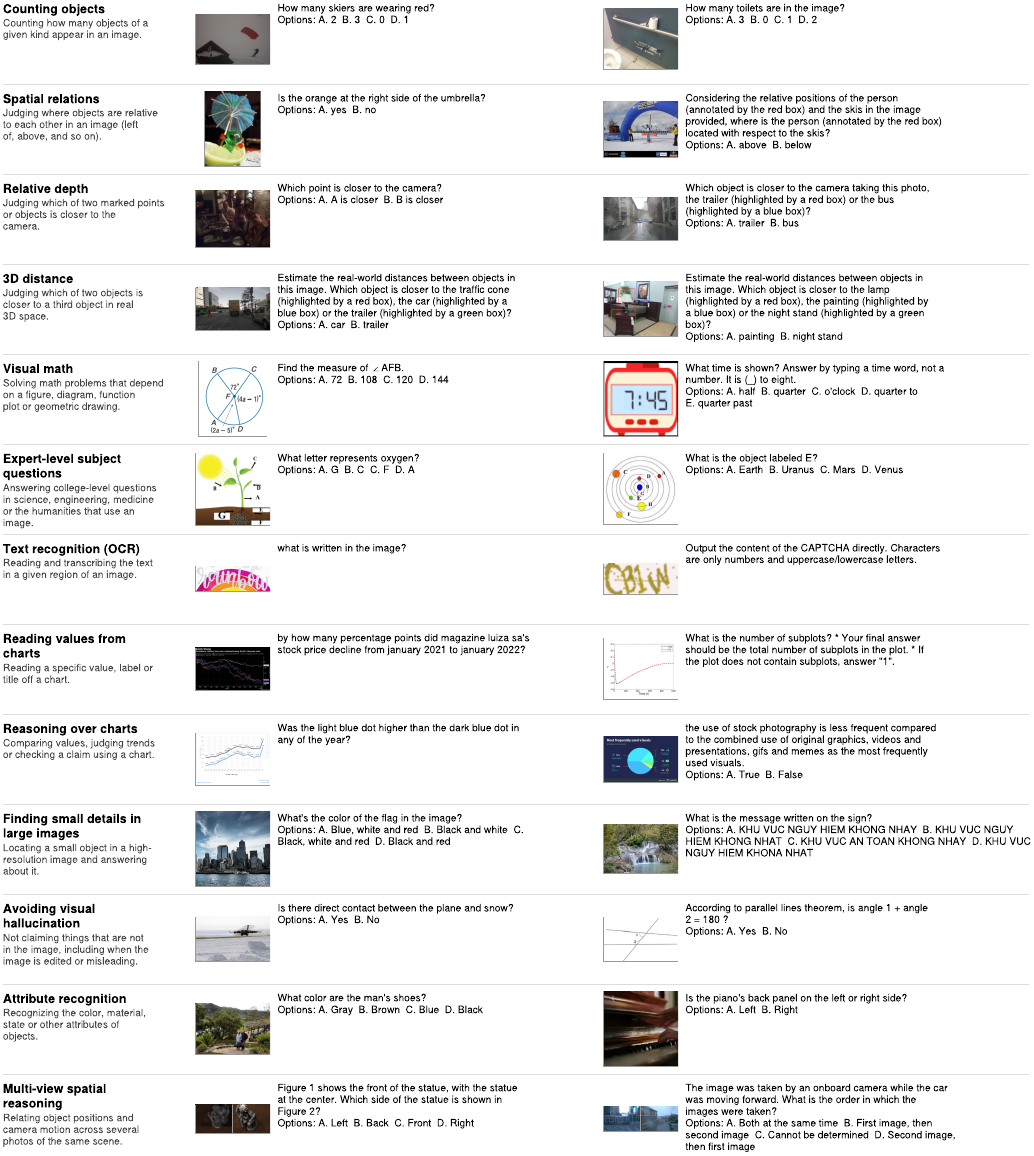}
  \caption{The name, definition, and two example questions of every axis, as shown in the prompts of \cref{sec:before-kinds}.
  The names follow the categories that model developers use in their own reports and differ slightly from those in the paper.}
  \label{fig:suppl-examples}
\end{figure*}
\begin{figure*}[t]
  \centering
  \includegraphics[width=\textwidth]{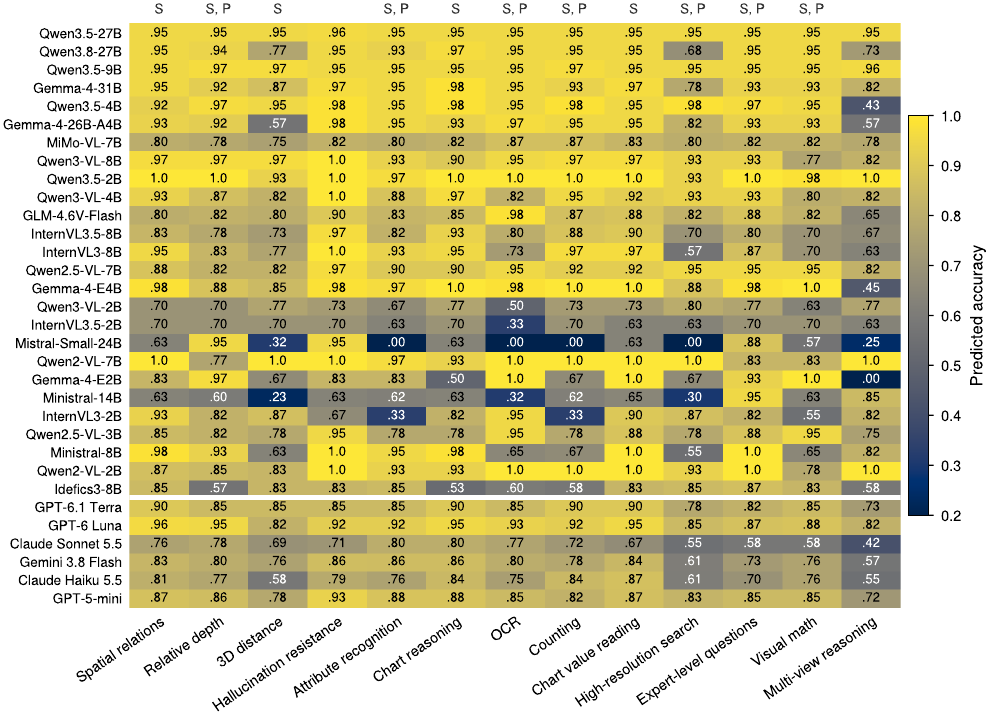}
  \caption{Every model's prediction on every axis with the single-axis prompt, averaged over the three wordings, in the layout of \cref{fig:suppl-profiles} for comparison.
  GLM-4.1V-9B-Thinking fails the format rule and is not shown.}
  \label{fig:suppl-predictions}
\end{figure*}
\Cref{fig:suppl-prompts} shows the prompts of \cref{sec:before-kinds}.
Each axis is described by its name, a one-sentence definition, and two example questions with their images (\cref{fig:suppl-examples}).
The examples are probe questions, which are never used for evaluation.
For each axis we take the first two short questions in a fixed random order, so the examples are not chosen by hand.
We read the prediction from the last ``Accuracy: $N$'' in the reply and average it over the three wordings.
For direct choices, the two axes are labeled X and Y so that the labels do not clash with the option letters of the examples.
\Cref{fig:suppl-predictions} gives every model's predictions.

\subsection{The reference and the partial correlation}
\label{sec:suppl-partial}
\begin{table}[t]
\caption{The partial correlation of \cref{sec:before-kinds} worked through for GLM-4.6V-Flash. Accuracy and prediction are in percent. Ref.\ is the reference, the mean accuracy of the other open models, and Resid.\ the residual in percentage points from a least-squares line on the reference across the axes.}
\label{tab:suppl-partial}
\centering
\scriptsize
\setlength{\tabcolsep}{3pt}
\begin{tabular}{@{}lrrrrr@{}}
\toprule
 & \multicolumn{3}{c}{Accuracy} & \multicolumn{2}{c}{Prediction} \\
\cmidrule(lr){2-4}\cmidrule(l){5-6}
Axis & Model & Ref. & Resid. & Model & Resid. \\
\midrule
  High-resolution search & 76 & 66 & +8 & 82 & $-$4 \\
  OCR & 81 & 72 & +6 & 98 & +13 \\
  Counting & 77 & 71 & +3 & 87 & +1 \\
  3D distance & 87 & 80 & +2 & 80 & $-$4 \\
  Relative depth & 89 & 84 & 0 & 82 & $-$1 \\
  Chart reasoning & 76 & 73 & $-$1 & 85 & 0 \\
  Expert-level questions & 57 & 58 & $-$2 & 88 & +1 \\
  Chart value reading & 69 & 69 & $-$3 & 88 & +3 \\
  Attribute recognition & 76 & 76 & $-$4 & 83 & $-$1 \\
  Spatial relations & 88 & 86 & $-$4 & 80 & $-$3 \\
  Visual math & 52 & 57 & $-$5 & 82 & $-$6 \\
\bottomrule
\end{tabular}
\end{table}

\Cref{sec:before-kinds} separates what a model's predictions say about difficulty shared by all models from what they say about the model itself.

\paragraph{Reference}
For a model, the reference on an axis is the mean accuracy of the open models other than the model itself on that axis.
A proprietary model is not among them, so its reference is the mean of all \nOpenVLMs{} open models.
Adding the six proprietary models to the reference leaves the conclusions unchanged.
The order of the seven compared axes stays the same, so no pair of axes changes, and every result for the proprietary models in \cref{sec:before-image,sec:where} keeps its sign.

\paragraph{Partial correlation}
Let $y_a$ be the model's accuracy on axis $a$, $p_a$ its prediction, and $R_a$ the reference, for the \nAxesL{} scored axes.
We fit $y_a = \alpha + \beta R_a + e_a$ by least squares across the axes, so a positive residual $e_a$ means that the model does better on axis $a$ than the reference suggests.
We fit $p_a = \alpha' + \beta' R_a + d_a$ in the same way.
The score is the Pearson correlation between the residuals $d_a$ and $e_a$ across the axes.
The fit absorbs any constant and any multiple of $R_a$, so a prediction that gives every axis the same value, or one that only follows the reference, scores zero.
A simpler score, the correlation between prediction minus reference and accuracy minus reference, gives a positive value even to a prediction that is the same for every kind (+0.18 and +0.34 for the untrained Qwen3.5-9B and Gemma-4-31B on the kinds of \cref{sec:learning-weights}), because the reference appears in both differences.

\paragraph{Worked example}
\Cref{tab:suppl-partial} works this through for GLM-4.6V-Flash, the model of \cref{fig:teaser}a.
Its accuracy is well above the reference on high-resolution search and text recognition, and below it on visual math.
Its predictions rise most on text recognition and fall most on visual math, which agrees with its accuracy, but they fall on high-resolution search, where its accuracy is highest relative to the reference.
Across the \nAxesL{} axes, the two residual columns correlate at \GLMStatedOwn{}.

\paragraph{Other models' predictions}
The second standard asks whether the model's own predictions track its residuals better than predictions made without any knowledge of the model.
For each other open model, we take the predictions it made about itself, compute their residuals $d_a$ against the same reference $R_a$, and correlate them with the model's accuracy residuals $e_a$.
The control score of the model is the mean of these correlations over the other models.
Beliefs that all models share, such as rating text recognition as easy, raise the control score as much as the model's own score, so the difference between the two leaves them out.
For GLM-4.6V-Flash the control score is \GLMStatedCtl{}.
Over the \nSzero{} open models, the own score is \StatedOwn{}, the control score is \StatedCtl{}, and their difference is \StatedDiff{} [\StatedDiffLo, \StatedDiffHi].
Intervals are 95\% bootstrap intervals over models with 10{,}000 resamples.

\paragraph{Other sets of axes}
The \nAxesL{} scored axes are the main set of the paper, and the \nAxesS{} compared axes are used only for the pair score.
Computed on the compared axes instead, the partial correlation of the stated predictions varies with the examples, from +0.02 to +0.24 over the five example sets of \cref{sec:suppl-robust41}, and only with the main examples does it lie clearly above zero and above the control.
The answer probability of \cref{sec:where}, which does not depend on examples, reaches +0.31 [+0.15, +0.47] there.
On all \nAxes{} axes, the difference between own and control predictions is \StatedDiffThirteen{} [\StatedDiffThirteenLo, \StatedDiffThirteenHi], just above zero.

\paragraph{Scale}
\begin{table*}[t]
\caption{The partial correlation of \cref{sec:before-kinds} computed in four ways, as means over models with 95\% bootstrap intervals. Empirical logit: accuracies as $\mathrm{logit}((k+0.5)/(n+1))$ and predictions as the logit of the predicted accuracy. Ranks: predictions, accuracies, and the reference replaced by their ranks across the axes. Additive logistic model: a model ability and an axis difficulty fitted to the other models' answers, and the model's accuracy compared with what ability plus difficulty predicts. Own $-$ others subtracts the same score computed with the other models' predictions. Answer probability is that of \cref{sec:where}, over all questions of an axis or five per axis, and on the 31 kinds of tasks of \cref{sec:learning-weights}.}
\label{tab:suppl-scale}
\centering
\scriptsize
\setlength{\tabcolsep}{2.5pt}
\begin{tabular}{@{}lccccc@{}}
\toprule
 & \multicolumn{2}{c}{Stated predictions, 11 axes} & \multicolumn{2}{c}{Answer prob., 11 axes} & Answer prob., 31 kinds \\
\cmidrule(lr){2-3}\cmidrule(lr){4-5}\cmidrule(l){6-6}
Way & Own & Own $-$ others & All questions & Five per axis & Own $-$ others \\
\midrule
  Linear (main) & +0.12 [$-$0.03, +0.27] & +0.13 [$-$0.001, +0.26] & +0.36 [+0.25, +0.47] & +0.26 [+0.18, +0.34] & +0.36 [+0.27, +0.44] \\
  Empirical logit & +0.13 [$-$0.02, +0.27] & +0.12 [$-$0.01, +0.25] & +0.36 [+0.24, +0.48] & +0.25 [+0.16, +0.34] & +0.38 [+0.30, +0.46] \\
  Ranks & +0.08 [$-$0.06, +0.21] & +0.09 [$-$0.02, +0.21] & +0.31 [+0.18, +0.42] & +0.20 [+0.12, +0.28] & +0.32 [+0.24, +0.41] \\
  Additive logistic & +0.12 [$-$0.01, +0.25] & +0.12 [$-$0.004, +0.24] & +0.33 [+0.22, +0.44] & +0.23 [+0.15, +0.31] & +0.36 [+0.28, +0.44] \\
\bottomrule
\end{tabular}
\end{table*}

Accuracy varies less near 0 and 1 than in the middle, so a linear fit could mistake a ceiling for a strength.
\Cref{tab:suppl-scale} therefore computes the partial correlation in three more ways: on an empirical logit scale, on ranks, and with an additive logistic model in which each model has an ability and each axis a difficulty.
In all four ways, the own stated predictions exceed the other models' by 0.09 to 0.13, with intervals that include zero, and the answer probability of \cref{sec:where} stays clearly above zero.

\subsection{Robustness}
\label{sec:suppl-robust41}
\begin{table*}[t]
\caption{The measures of \cref{sec:before-kinds} for other ways of asking. Pairs is the percentage of pairs of compared axes in which the predictions pick the harder axis for the model (chance 50, the other models' average 94), and Above is the number of models that do at least as well as the other models' average on the same pairs. Part.\ is the partial correlation with a 95\% bootstrap interval, and Own $-$ others subtracts the same score computed with the other models' predictions. Over.\ is the mean predicted minus actual accuracy over the scored axes, in points.}
\label{tab:suppl-robust}
\centering
\scriptsize
\setlength{\tabcolsep}{4pt}
\begin{tabular}{@{}lrrccr@{}}
\toprule
Prompt & Pairs & Above & Part. & Own $-$ others & Over. \\
\midrule
  Main: examples 1--2 & 52 & 0 & +0.12 [$-$0.03, +0.28] & +0.13 [$-$0.002, +0.26] & +11.5 \\
  Examples 3--4 & 62 & 0 & +0.02 [$-$0.10, +0.14] & $-$0.002 [$-$0.13, +0.13] & +13.1 \\
  Examples answered right twice & 68 & 2 & +0.04 [$-$0.08, +0.17] & +0.02 [$-$0.12, +0.15] & +12.4 \\
  Examples answered right once & 58 & 1 & +0.02 [$-$0.10, +0.15] & +0.002 [$-$0.13, +0.14] & +12.1 \\
  Examples answered wrong twice & 60 & 1 & +0.05 [$-$0.09, +0.19] & +0.04 [$-$0.10, +0.17] & +12.1 \\
\midrule
  All 13 axes in one prompt & 62 & 0 & $-$0.06 [$-$0.20, +0.08] & $-$0.05 [$-$0.12, +0.02] & +14.1 \\
  Answers of 0\% as missing & 52 & 0 & +0.08 [$-$0.07, +0.24] & +0.11 [$-$0.03, +0.24] & +16.4 \\
  Mistral models left out (23 models) & 51 & 0 & +0.06 [$-$0.10, +0.21] & +0.09 [$-$0.05, +0.24] & +13.8 \\
\bottomrule
\end{tabular}
\end{table*}

\Cref{tab:suppl-robust} repeats the measures of \cref{sec:before-kinds} for other ways of asking.

\paragraph{Examples}
We replaced the two examples of every axis by the next two probe questions that pass the example rule, and by two that the model itself had answered correctly twice, once, or not at all.
The mean prediction is the same in all three cases (84.5\%, 84.2\%, and 84.2\%), so the predictions do not depend on whether the model answers its examples correctly.
With the other example sets, the predictions pick the harder axis more often than with the main examples (58--68\% of pairs against 52\%), but still far less often than the other models' average (94\%), which at most two models reach.
Their partial correlation is smaller (0.02 to 0.05 against 0.12).

\paragraph{All axes at once}
Listing all 13 axes in one prompt, each with its definition and two examples as text, and asking for every axis in one reply gives the same picture, with 62\% of pairs and a partial correlation of $-0.06$.

\paragraph{Answers of 0\%}
Some models answer 0\% on many axes, mostly the three Mistral models, often with a sentence saying that they cannot process images.
Treating these answers as missing, or leaving out the three Mistral models, changes neither the pair score nor the partial correlation much.

\paragraph{Reference from other models}
Built instead from random subsets of 5, 8, or 13 other models, from models of other families only, or from the eight weakest or the eight strongest models, the reference still predicts a model's accuracy per axis better than the model's own predictions do for at least 20 of the 26 models.

\paragraph{A typical VLM}
We also asked the single-axis prompt about ``a typical vision-language model'' instead of the model itself.
These answers follow the general difficulty of the axes in 60\% of pairs, against 46\% when the models are asked about themselves, and even pick the harder axis for the model itself more often (59\% against 52\%).
The models thus know a little about which kinds of tasks are hard in general but do not apply it to themselves.

\paragraph{Perception and reasoning}
Half of the variance of the model-by-axis part (\cref{sec:suppl-axes}) lies along one direction, from perception axes (high-resolution search, OCR, counting, and attribute recognition) to reasoning axes (visual math, chart value reading, and expert-level questions).
Only 14 of the 26 models predict correctly whether they lean further toward perception or toward reasoning than the other models do, about as many as chance would give.
On average, their predictions put the reasoning axes 7 points above the perception axes, whereas their accuracy is 10 points higher on the perception axes.
The correlation between predicted and actual gaps, +0.38, falls to +0.14 without the Mistral models' answers of 0\%.
The models thus do not know even the main direction in which their strengths differ from those of other models.

\subsection{Direct choice per pair}
\label{sec:suppl-pairs}
\begin{figure}[t]
  \centering
  \includegraphics[width=\linewidth]{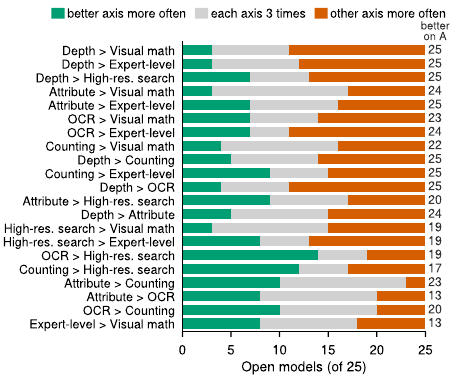}
  \caption{Choosing directly between two axes, for all 21 pairs of compared axes.
  Each pair is written $A > B$, with $A$ the axis of higher mean accuracy over the models, and pairs are ordered by this gap.
  For each pair, the bars count the open models that mostly choose the axis they are more accurate on, split their six answers evenly, or mostly choose the other axis, and the right column gives the number of models that are more accurate on $A$.}
  \label{fig:suppl-pairs}
\end{figure}
Each pair of compared axes is asked in both orders and with three wordings, so each model gives six answers per pair (\cref{fig:suppl-pairs}).
In \PairsWorseMost{} of the 21 pairs, and in all \PairsWorseTop{} pairs with the largest gaps, more models choose the axis they are worse at.
An even split usually comes from choosing the axis shown first in both orders, and \FirstShownPct\% of all answers name the axis shown first.
With the examples as text only, the choices are again at chance (45\% [40, 51] of the weighted pairs), and no model reaches the other models' average.

\section{Predictions under image conditions}
\label{sec:suppl-image}
\begin{figure}[t]
  \centering
  \includegraphics[width=0.875\linewidth]{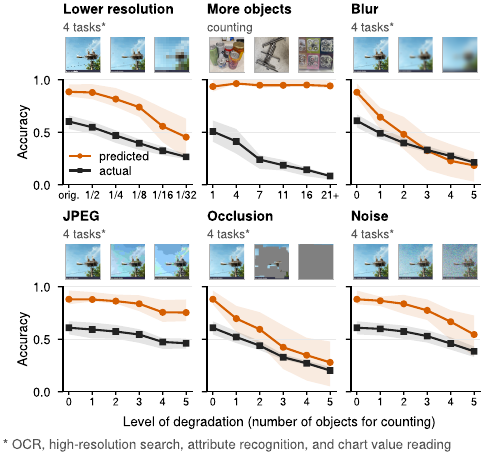}
  \caption{Predicted and actual accuracy under image conditions, as in \cref{fig:image} but averaged over the open models, with the middle half of the models shaded.}
  \label{fig:suppl-image-mean}
\end{figure}
\begin{table}[t]
\caption{Number of open models with each kind of predicted drop under each image condition, the counts that \cref{sec:before-image} reports for some of the conditions. Worse counts the models whose predicted drop is further from their actual drop than the reference is, Fall those that predict a fall of at least 10 points from the first to the last level, Flat those whose prediction stays within 5 points, and Larger those that predict a larger drop than they show.}
\label{tab:suppl-image}
\centering
\scriptsize
\setlength{\tabcolsep}{3pt}
\begin{tabular}{@{}lrrrrr@{}}
\toprule
Condition & Models & Worse & Fall & Flat & Larger \\
\midrule
  Lower resolution & 27 & 26 & 24 & 3 & 17 \\
  More objects & 27 & 23 & 0 & 22 & 0 \\
  Blur & 25 & 25 & 24 & 0 & 23 \\
  JPEG compression & 25 & 24 & 12 & 8 & 9 \\
  Occlusion & 25 & 25 & 23 & 1 & 18 \\
  Noise & 25 & 25 & 22 & 1 & 13 \\
\bottomrule
\end{tabular}
\end{table}

\begin{figure}[t]
  \centering
  \includegraphics[width=\linewidth]{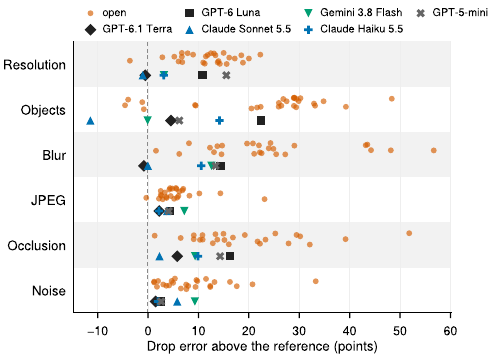}
  \caption{How much further each model's predicted drop is from its actual drop than the reference is, in points, averaged over the levels and axes of each image condition.
  Above 0, the other models' average drop predicts the model's drop better than the model does.
  Dots are open models and the other markers proprietary models.}
  \label{fig:suppl-image-excess}
\end{figure}
\paragraph{Levels}
For lower resolution, each side of the image is scaled to 1, 1/2, 1/4, 1/8, 1/16, and 1/32 of its original length, and the image is not scaled back up.
Blur is Gaussian with a standard deviation of 0.4, 0.8, 1.6, 3.2, and 6.4\% of the shorter side, JPEG compression uses qualities 20, 10, 6, 3, and 1, occlusion covers 20, 35, 65, 80, and 95\% of the image with grey rectangles, and noise is Gaussian with a standard deviation of 8, 15, 25, 40, and 60\% of the intensity range.
We chose these five levels in a pilot with three models so that accuracy falls in roughly even steps from the original toward chance.
For object counts, the questions come from CountQA\footnote{J. S. Tamarapalli, R. Grover, N. Pande, and S. Yerramilli. CountQA: How Well Do MLLMs Count in the Wild? arXiv:2508.06585, 2025.}, real photos grouped by their true count into 1--3, 4--6, 7--10, 11--15, 16--20, and 21--50 objects.
Each level is scored on 100 evaluation questions per axis, the same questions at every level of a condition.
An image below the smallest size a model accepts counts as answered wrong.

\paragraph{Predictions}
The prompt shows three example questions of the axis with their images at the level and asks for the accuracy on 100 such questions (\cref{sec:before-image}).
We ask with three sets of three example questions, all probe questions, and with three wordings, and average the nine answers.
The drop error is the mean absolute difference between the predicted and the actual drop from the first level, over the five levels after it.
The reference uses the other open models' mean actual drop in place of the predicted drop.

\paragraph{Results}
\Cref{fig:suppl-image-mean} shows the curves averaged over the open models, \cref{tab:suppl-image} counts the models condition by condition, and \cref{fig:suppl-image-excess} shows every model.
In every condition, the reference predicts the drop of nearly every model better than the model does, and over the four degradations together it does so for all \nHsixc{} models.
Under JPEG compression, both the predicted and the actual drops are small, so the difference is smallest there.
Among the proprietary models, GPT-6.1 Terra and Claude Sonnet 5.5 come closest to the reference.
On lower resolution they match it, with predicted drops 0.5 and 1.1 points closer to the actual drops than the reference is, but the reference fits them poorly, since their accuracy starts about 20 points higher than that of the open models and falls to the same level, so it misses their drops by 13.5 and 12.1 points against 6.4 for the median open model.
On the four degradations they stay 2 and 3 points behind it.
On counting, Claude Sonnet 5.5 predicts its drop much better than the reference and Gemini 3.8 Flash about as well, because their accuracy falls far less than that of the open models as objects become more numerous.

\section{Records in context}
\label{sec:suppl-record}
\begin{figure}[t]
  \centering\scriptsize
  \fbox{\begin{minipage}{0.95\linewidth}\ttfamily\raggedright
  [the single-axis prompt of \cref{fig:suppl-prompts}a up to the examples]\\[3pt]
  Here is a record of questions of this kind that you answered earlier, in random order. It shows your answers and whether each one was correct.\\
  {[1]} Question: How many motorcycles are in the image?\ \ Options: A. 0\ \ B. 2\ \ C. 1\ \ D. 3\\
  \quad Your answer: C (1)\\
  \quad Result: correct\\
  {[2]} Question: How many light bulbs are in the picture?\\
  \quad Your answer: 2\\
  \quad Result: incorrect (correct answer: 3)\\[3pt]
  If you were given 100 new questions of this kind, each with its image, what percentage would you answer correctly?\\
  Reply with exactly: "Accuracy: <number from 0 to 100>"
  \end{minipage}}
  \caption{The prompt with a record of two questions, for counting, with the record of Qwen3-VL-8B.
  Without correctness, the second sentence of the record reads ``It shows your answers but not whether they were right.'' and the Result lines are left out.
  The count line replaces the record by ``Here is a summary of questions of this kind that you answered earlier.'' followed by ``Counting objects: 1 of 2 correct (50\%)''.}
  \label{fig:suppl-record-prompt}
\end{figure}
\begin{table}[t]
\caption{Mean error of predicted accuracy over the open models, in points, by the number of questions per axis in the record. The record shows the answers with or without whether they were correct, or only a line with the count of correct answers. The last two columns are not asked: the accuracy in the record used as the prediction, and the best estimate from the record, which moves the accuracy in the record toward the other models' average accuracy, the more so the shorter the record.}
\label{tab:suppl-record}
\centering
\scriptsize
\setlength{\tabcolsep}{3pt}
\begin{tabular}{@{}lrrrrr@{}}
\toprule
 & \multicolumn{3}{c}{Asked} & \multicolumn{2}{c}{Not asked} \\
\cmidrule(lr){2-4}\cmidrule(l){5-6}
Record & \shortstack{With\\correctness} & \shortstack{Answers\\only} & \shortstack{Count\\line} & \shortstack{Accuracy\\in record} & \shortstack{Best\\estimate} \\
\midrule
  None & 22.0 & & & & \\
  2 & 20.6 & 22.8 & 23.4 & 25.4 & 7.5 \\
  5 & 15.5 & 20.4 & 16.7 & 16.3 & 7.0 \\
  15 & 12.3 & 18.8 & 11.8 & 8.7 & 5.8 \\
\bottomrule
\end{tabular}
\end{table}

\begin{figure}[t]
  \centering
  \includegraphics[width=\linewidth]{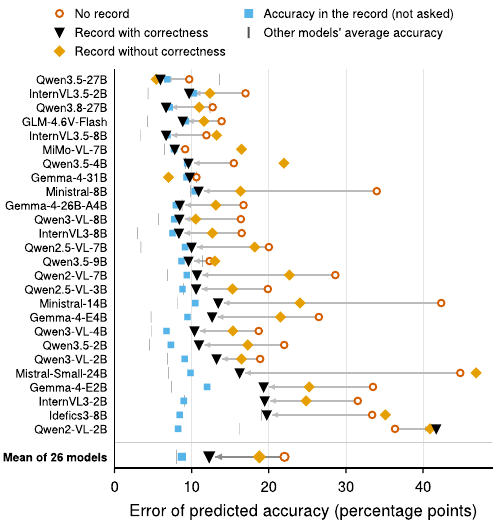}
  \caption{\Cref{fig:record}a for all \nSzero{} open models: error of predicted accuracy with 15 questions per axis in the record.
  Models are ordered by how close the record with correctness brings them to the accuracy in the record, and short marks show the other models' average accuracy used as the prediction.}
  \label{fig:suppl-record-models}
\end{figure}
\paragraph{Records}
The record of a model on an axis consists of probe questions that the model answered earlier.
For each length we draw five different records, ask each with the three wordings, and average the fifteen answers.
Each entry gives the question, the model's answer, and whether it was correct (\cref{fig:suppl-record-prompt}).
We also show the same entries without correctness, a single line with the count of correct answers, another model's entries presented as the model's own, and, for two questions per axis, the entries with their images.

\paragraph{Results}
\Cref{tab:suppl-record} gives the error by record length, and \cref{fig:suppl-record-models} gives every model.
Replacing the entries by the count of correct answers leaves the error about the same (\RecCount{} against \RecErrFifteen{} points with 15 questions), so counting the correct answers is not what limits the models.
With the count given, \RecHigherPct\% of the answers that differ from the shown accuracy lie above it, so part of the error is overestimation that remains even when the model is told how often it was right.
When another model's record is presented as the model's own, the prediction moves \RecSlopeTen{} points [\RecSlopeTenLo, \RecSlopeTenHi] for every 10 points of difference between the two records, and it moves toward the other record for all \RecSlopePos{} models, so the models follow the evidence they are shown, whoever it comes from.
With two questions per axis, adding the images to the entries changes the error by less than a point (21.3 against 20.6 points).
The best estimate from the same record would have an error of 5.8 points with 15 questions, about half that of the models.
Showing the records of all 13 axes mixed together, without axis names, helps more than the record of the asked axis when the records are short (14.9 against 20.6 points with two questions per axis), since the other axes also show how often the model is right overall, and slightly less with 15 questions (13.6 against 12.3).

\section{Training on past outcomes}
\label{sec:suppl-train}
\begin{table*}[t]
\caption{The 31 kinds of tasks of \cref{sec:learning-weights} by their names in the prompts, grouped by axis and by the fold that leaves the axis out. Each fold also leaves out one degradation of the image conditions (fold 1 blur; fold 2 JPEG compression; fold 3 occlusion; fold 4 noise), and the object counts go with counting.}
\label{tab:suppl-kinds}
\centering
\scriptsize
\setlength{\tabcolsep}{3pt}
\begin{tabularx}{\textwidth}{@{}c>{\raggedright\arraybackslash}p{2.6cm}>{\raggedright\arraybackslash}X@{}}
\toprule
Fold & Axis & Kinds of tasks (source) \\
\midrule
  1 & OCR & Reading a word in a small crop (OCRBench, recognition); Reading text in photos, CAPTCHAs and regions (OCRBench v2, recognition) \\
  1 & Relative depth & Which boxed object is closer (CV-Bench, depth); Which marked point is closer (BLINK, relative depth) \\
  1 & Spatial relations & Left/right or above/below of a boxed object (CV-Bench, relation); Checking a statement about placement (BLINK, spatial relation) \\
  \addlinespace[2pt]
  2 & High-resolution search & Small details in 4K photos (HR-Bench, 4K); Small details in 8K photos (HR-Bench, 8K); Color or side of a small object (V*Bench) \\
  2 & 3D distance & 3D distance (CV-Bench, distance) \\
  2 & Hallucination resistance & Is it really in the photo (AMBER); Edited or misleading images (HallusionBench, image questions) \\
  2 & Multi-view reasoning & Multi-view spatial reasoning (MMSI-Bench) \\
  \addlinespace[2pt]
  3 & Attribute recognition & Attributes in everyday photos (SEED-Bench, instance attributes); Easily overlooked visual details (MMVP); Short questions on specific details (MMStar, fine-grained perception) \\
  3 & Counting & Counting one kind of object (CV-Bench, count); Open-ended counting (TallyQA); Counting objects that meet a condition (BLINK, counting) \\
  3 & Expert-level questions & College exam questions with images (MMMU-Pro, standard, four options); School-level diagram and science questions (MMStar, reasoning) \\
  \addlinespace[2pt]
  4 & Chart value reading & Facts from real-world charts (ChartQAPro, factoid); Yes/no checks on scientific plots (CharXiv, descriptive, yes/no); Counts and tick values in scientific plots (CharXiv, descriptive, numeric); Labels in scientific plots (CharXiv, descriptive, text) \\
  4 & Chart reasoning & Questions on simple charts (ChartQA, human-written); Checking claims against a chart (ChartQAPro, fact-checking); Multi-step chart calculations (ChartQAPro, multiple choice) \\
  4 & Visual math & Geometry read from the figure (MathVerse, vision-dominant); Math competition problems (MATH-Vision, test); Varied math in pictures (MathVista, multiple choice) \\
\bottomrule
\end{tabularx}
\end{table*}

\begin{table}[t]
\caption{Candidates for the other model of \cref{sec:learning-weights}: the four open models within 10 points of the trained model's accuracy whose strengths and weaknesses relative to the other models are least like its own, measured by the partial correlation of the two models' accuracies on kinds of tasks and on image conditions. * marks the model chosen, the least alike among those answered with the same inference settings; $^\dagger$ marks models left out because their outputs differ between the two inference engines.}
\label{tab:suppl-donor}
\centering
\scriptsize
\setlength{\tabcolsep}{3pt}
\begin{tabular}{@{}llrrr@{}}
\toprule
Trained model & Candidate & Kinds & Conditions & Mean \\
\midrule
  Qwen3.5-9B & InternVL3.5-8B$^\dagger$ & $-$0.21 & $-$0.17 & $-$0.19 \\
   & GLM-4.6V-Flash* & +0.02 & +0.03 & +0.02 \\
   & Qwen3-VL-4B & +0.17 & +0.01 & +0.09 \\
   & Gemma-4-26B-A4B & +0.33 & +0.22 & +0.27 \\
  \addlinespace[2pt]
  Gemma-4-31B & GLM-4.6V-Flash* & $-$0.44 & $-$0.30 & $-$0.37 \\
   & Qwen3-VL-4B & $-$0.29 & $-$0.38 & $-$0.34 \\
   & Qwen3.5-2B & +0.04 & $-$0.21 & $-$0.09 \\
   & InternVL3-8B$^\dagger$ & $-$0.06 & +0.17 & +0.06 \\
  \addlinespace[2pt]
  GLM-4.6V-Flash & Gemma-4-26B-A4B* & $-$0.48 & $-$0.48 & $-$0.48 \\
   & Gemma-4-E4B & $-$0.50 & $-$0.30 & $-$0.40 \\
   & Mistral-Small-24B$^\dagger$ & $-$0.46 & $-$0.33 & $-$0.40 \\
   & Gemma-4-31B & $-$0.45 & $-$0.33 & $-$0.39 \\
\bottomrule
\end{tabular}
\end{table}

\begin{table}[t]
\caption{\Cref{tab:train} with the last row of each model trained on the accuracy of a second other model, Qwen3.5-9B or Gemma-4-31B. The other rows and the columns are those of \cref{tab:train}.}
\label{tab:suppl-train}
\centering
\footnotesize
\setlength{\tabcolsep}{3pt}
\resizebox{\linewidth}{!}{%
\begin{tabular}{@{}lrrrrrr@{}}
\toprule
 & \multicolumn{3}{c}{Kinds of tasks} & \multicolumn{3}{c}{Image conditions} \\
\cmidrule(lr){2-4}\cmidrule(l){5-7}
Trained on & Over.$\to$0 & Pairs$\uparrow$ & Part.$\uparrow$ & Over.$\to$0 & Pairs$\uparrow$ & Part.$\uparrow$ \\
\midrule
  \multicolumn{7}{@{}l}{\textit{Qwen3.5-9B}} \\
  \quad Untrained & +15.6 & 48\,/\,48 & +0.08 & +32.9 & 67\,/\,62 & +0.45 \\
  \quad Own & +2.3 & 53\,/\,42 & $-$0.49 & +7.3 & 68\,/\,61 & +0.49 \\
  \quad Gemma-4-31B & +2.3 & 47\,/\,45 & $-$0.26 & +7.6 & 73\,/\,59 & +0.42 \\
\addlinespace[3pt]
  \multicolumn{7}{@{}l}{\textit{Gemma-4-31B}} \\
  \quad Untrained & +12.8 & 57\,/\,25 & +0.12 & +14.6 & 68\,/\,58 & +0.24 \\
  \quad Own & $-$0.9 & 48\,/\,39 & $-$0.11 & +8.9 & 70\,/\,64 & $-$0.02 \\
  \quad Qwen3.5-9B & +4.0 & 51\,/\,43 & $-$0.38 & +7.3 & 67\,/\,61 & +0.01 \\
\addlinespace[3pt]
  \multicolumn{7}{@{}l}{\textit{GLM-4.6V-Flash}} \\
  \quad Untrained & +14.8 & 50\,/\,57 & $-$0.01 & +18.9 & 59\,/\,61 & $-$0.02 \\
  \quad Own & +0.2 & 68\,/\,63 & +0.36 & +3.2 & 66\,/\,63 & +0.47 \\
  \quad Qwen3.5-9B & +11.5 & 61\,/\,47 & +0.39 & +12.0 & 65\,/\,66 & +0.48 \\
\midrule
  Other models' average & & 94--96\,/\,0 & & & 98\,/\,0 & \\
\bottomrule
\end{tabular}}
\end{table}

\begin{table*}[t]
\caption{Differences in the partial correlation between the model trained on its own accuracy and the model named in each column, which is the other model of \cref{tab:train}, the second other model of \cref{tab:suppl-train}, or the untrained model, with 95\% bootstrap intervals.}
\label{tab:suppl-trainci}
\centering
\scriptsize
\setlength{\tabcolsep}{2.5pt}
\resizebox{\textwidth}{!}{%
\begin{tabular}{@{}lcccccc@{}}
\toprule
 & \multicolumn{3}{c}{Kinds of tasks} & \multicolumn{3}{c}{Image conditions} \\
\cmidrule(lr){2-4}\cmidrule(l){5-7}
Model & Other & Second other & Untrained & Other & Second other & Untrained \\
\midrule
  Qwen3.5-9B & $-$0.18 [$-$0.49, +0.14] & $-$0.23 [$-$0.55, +0.10] & $-$0.57 [$-$0.87, $-$0.25] & +0.13 [$-$0.03, +0.31] & +0.07 [$-$0.03, +0.16] & +0.04 [$-$0.19, +0.24] \\
  Gemma-4-31B & +0.01 [$-$0.49, +0.84] & +0.27 [$-$0.21, +0.65] & $-$0.23 [$-$0.78, +0.78] & +0.07 [$-$0.23, +0.36] & $-$0.03 [$-$0.28, +0.25] & $-$0.27 [$-$0.44, $-$0.06] \\
  GLM-4.6V-Flash & $-$0.03 [$-$0.44, +0.39] & $-$0.02 [$-$0.43, +0.38] & +0.37 [$-$0.17, +0.76] & +0.12 [$-$0.09, +0.41] & $-$0.01 [$-$0.10, +0.12] & +0.49 [+0.12, +0.80] \\
\bottomrule
\end{tabular}}
\end{table*}

\paragraph{Kinds and conditions}
The 31 kinds of tasks split the axes by source benchmark or benchmark subset (\cref{tab:suppl-kinds}).
Each kind has its own name and definition, written from its questions, so that the kinds of one axis differ in more than their examples.
The 109 image conditions are the six resolutions of the four axes of \cref{sec:before-image}, the five levels of each degradation on the same axes, and five of the six ranges of object counts, since the range of 1--3 objects has no questions left for training.

\paragraph{Folds}
Each fold leaves out the axes of \cref{tab:suppl-kinds} with all their kinds and conditions, and one degradation, so that between 21 and 25 kinds of tasks remain for training.
A kind or condition is evaluated with the model of the first fold that leaves it out.

\paragraph{Training prompts}
The training set of a fold, about \TrainRowsPerFold{} prompts, has four equal parts.
Two ask for the accuracy on a kind of task or an image condition with the prompts of \cref{sec:before-kinds,sec:before-image}, and two ask with direct-choice prompts which of two kinds or two conditions the model does better on.
The answer is the model's accuracy on the probe questions of the kind or the training questions of the condition, and for a pair the one with the higher accuracy, using only pairs that differ by at least 10 points.
The example questions of the training prompts never appear in the evaluation prompts.

\paragraph{Training}
We train LoRA adapters of rank 16, with scale 32 and dropout 0.05, on all attention and MLP projections of the language model, with AdamW at a learning rate of $10^{-4}$, a linear warm-up over the first 5\% of the steps and a linear decay to zero, gradients accumulated over 8 prompts, and two epochs.
The loss covers only the answer, and the images keep their original size.
The trained adapters are evaluated in vLLM, like the untrained versions of these models (\cref{sec:suppl-axes}).
We use supervised fine-tuning because the model's past accuracy on each kind is a known target.
Other training methods might do better, since evolution strategies teach language models to judge whether they know the answer to a single question better than supervised fine-tuning does~\cite{esma}.

\paragraph{Pairs}
The Pairs score of \cref{tab:train} uses every pair of kinds of tasks left out by the same fold, so that the model was trained on neither kind, which gives 15, 21, 28, and 45 pairs in the four folds.
For image conditions it uses 200 sampled pairs per fold.
Each pair is asked in both orders with the three wordings.

\paragraph{The other model}
Besides the rule of \cref{sec:learning-weights}, the other model must be one whose answers come out the same through the inference engine used for training, which leaves out the InternVL and Mistral models (\cref{tab:suppl-donor}).
We also trained each model on the accuracy of a second other model, Qwen3.5-9B or Gemma-4-31B (\cref{tab:suppl-train}), with the same conclusions.

\paragraph{Intervals}
\Cref{tab:suppl-trainci} gives the differences in the partial correlation with 95\% bootstrap intervals that resample the axes for kinds of tasks and the degradation curves for image conditions, since the kinds of one axis and the levels of one curve are not independent.
For no model and neither type does training on its own accuracy give a reliably larger partial correlation than training on another model's, and the rise of GLM-4.6V-Flash with training is reliable only on image conditions.
Scoring the same predictions on the 11 axes of \cref{sec:before-kinds} leads to the same conclusions, with wider intervals over the fewer units.

\section{Decisions}
\label{sec:suppl-decisions}
\begin{table*}[t]
\caption{Values of \cref{fig:decisions} and of the threshold decision, before and after training. Each cell gives Qwen3.5-9B\,/\,Gemma-4-31B\,/\,GLM-4.6V-Flash. The other model is the one of \cref{sec:learning-weights}, and for the image budget the model whose drops are least like those of the trained model. Handing off uses the 11 axes and the questions of \cref{sec:before-kinds}, and the threshold uses the kinds of \cref{tab:train}, with the threshold at the model's median accuracy over the kinds or averaged over fixed thresholds from 0.5 to 0.9. The image budget is decided by models trained on drops.}
\label{tab:suppl-decisions}
\centering
\footnotesize
\setlength{\tabcolsep}{3pt}
\begin{tabular}{@{}lcccc@{}}
\toprule
 & & \multicolumn{2}{c}{Trained on the outcomes of} & Other models' \\
\cmidrule(lr){3-4}
Decision & Untrained & itself & another model & average \\
\midrule
Handing off$\uparrow$ & 0.62\,/\,0.16\,/\,$-$0.48 & 0.74\,/\,0.29\,/\,0.74 & 0.70\,/\,0.70\,/\,0.06 & 0.89\,/\,0.84\,/\,0.78 \\
Threshold at median, kinds$\uparrow$ & 0.00\,/\,0.26\,/\,0.20 & 0.50\,/\,0.34\,/\,0.78 & 0.60\,/\,0.45\,/\,0.02 & 0.77\,/\,0.70\,/\,0.95 \\
Threshold at median, conditions$\uparrow$ & 0.39\,/\,0.58\,/\,0.57 & 0.57\,/\,0.54\,/\,0.57 & 0.50\,/\,0.32\,/\,0.52 & 0.92\,/\,0.91\,/\,0.95 \\
Threshold 0.5--0.9, kinds$\uparrow$ & 0.00\,/\,0.08\,/\,0.32 & 0.10\,/\,0.10\,/\,0.25 & $-$0.21\,/\,0.20\,/\,0.32 & 0.10\,/\,$-$0.20\,/\,0.92 \\
Threshold 0.5--0.9, conditions$\uparrow$ & 0.43\,/\,0.61\,/\,0.65 & 0.80\,/\,0.75\,/\,0.86 & 0.83\,/\,0.76\,/\,0.85 & 0.94\,/\,0.94\,/\,0.99 \\
Image budget (levels)$\downarrow$ & 1.5\,/\,1.2\,/\,2.3 & 0.9\,/\,1.0\,/\,1.0 & 1.5\,/\,0.8\,/\,1.3 & 0.6\,/\,1.0\,/\,0.5 \\
\bottomrule
\end{tabular}
\end{table*}

\paragraph{Handing off}
Handing off is scored as in \cref{sec:learning-decisions}, on 50 evaluation questions from each of the 11 scored axes.
Before training, the scores of the \nSzero{} open models range from $-0.81$ to $0.94$, with a median of 0.00, and only one model reaches the other models' average.
The three trained models are scored on the same questions before and after training (\cref{tab:suppl-decisions}).

\paragraph{Answering above a threshold}
A model answers the questions of a kind if its predicted accuracy on the kind reaches a threshold and hands them off otherwise.
An answered kind gains its accuracy minus the threshold, a handed-off kind gains nothing, and the total is scaled so that answering every kind gives 0 and deciding with the actual accuracy gives 1.
\Cref{tab:suppl-decisions} gives the results.
With the threshold at the model's median accuracy over the kinds, training on its own accuracy raises the score on kinds of tasks for all three models, most for GLM-4.6V-Flash (from 0.20 to 0.78, against 0.02 when trained on another model's), while training on another model's accuracy raises it more for Qwen3.5-9B and Gemma-4-31B.
On image conditions, training on its own accuracy raises the score only for Qwen3.5-9B.
Averaged over fixed thresholds from 0.5 to 0.9, training raises it on image conditions instead, as much with another model's accuracy as with its own, and changes it little on kinds of tasks.
The other models' average accuracy scores at least as high in all but one of the twelve comparisons, and because the result depends on the threshold, the main text reports the two decisions that do not need one.

\paragraph{Image budget}
An image costs its number of pixels under lower resolution and its file size under JPEG compression, and the tolerated losses go from 0 to 20 points in steps of 1 point.
The distance from the best choice is averaged over the eight curves and the tolerated losses.
With the drops implied by their predictions in \cref{sec:before-image}, 24 of the 25 open models choose worse than they would with the other models' average drops, and they lose 7.8 points of accuracy against the best choice on average, against 1.0 point with the other models' drops.

\paragraph{Training to predict the drop}
For training on drops, the prompt shows an image at its original quality and the same image at one level of a degradation, with the question asked about it, and asks how many points lower the model's accuracy would be with such images.
The kinds are the five levels of resolution, blur, JPEG compression, occlusion, and noise on the four axes, 100 in all, with four folds by axis as in \cref{sec:suppl-train}.
The answer is the model's own drop on the training questions of the kind, from the original to the level, and the training settings are those of \cref{sec:suppl-train}.
The other model is chosen by the rule of \cref{sec:learning-weights} with the likeness of the two models' drops: Gemma-4-26B-A4B for Qwen3.5-9B, Qwen3-VL-8B for Gemma-4-31B, and Qwen3.5-4B for GLM-4.6V-Flash.
Training on its own drops brings the choices closer to the best than training on the other model's drops for Qwen3.5-9B and GLM-4.6V-Flash and less close for Gemma-4-31B (\cref{tab:suppl-decisions}), with intervals that include zero for all three.
Training on accuracy (\cref{sec:learning-weights}) does not help with this decision, since the models trained that way choose 1.7 levels from the best on average, the same as before training, against \DropOwn{} after training on drops.

\section{Probing the hidden states}
\label{sec:suppl-probe}
\begin{table*}[t]
\caption{Linear probes on hidden states before solving. Each row predicts one model's accuracy on held-out kinds from the hidden states of each of the three models (its own in bold) and gives the partial correlation of \cref{sec:before-kinds} with a 95\% bootstrap interval; the last column is the model's own stated prediction before training.}
\label{tab:suppl-probe}
\centering
\scriptsize
\setlength{\tabcolsep}{3pt}
\resizebox{\textwidth}{!}{%
\begin{tabular}{@{}llcccc@{}}
\toprule
 & & \multicolumn{3}{c}{Probe on the hidden states of} & \\
\cmidrule(lr){3-5}
Type & Accuracy of & Qwen3.5-9B & Gemma-4-31B & GLM-4.6V-Flash & Stated \\
\midrule
  Kinds of tasks & Qwen3.5-9B & \textbf{$-$0.38 [$-$0.66, $-$0.04]} & $-$0.03 [$-$0.46, +0.31] & $-$0.27 [$-$0.72, +0.32] & +0.08 [$-$0.13, +0.38] \\
   & Gemma-4-31B & +0.00 [$-$0.35, +0.43] & \textbf{$-$0.24 [$-$0.65, +0.31]} & $-$0.33 [$-$0.59, +0.24] & +0.12 [$-$0.46, +0.38] \\
   & GLM-4.6V-Flash & $-$0.13 [$-$0.50, +0.42] & +0.19 [$-$0.24, +0.57] & \textbf{+0.22 [$-$0.21, +0.57]} & $-$0.01 [$-$0.26, +0.40] \\
  \addlinespace[2pt]
  Image conditions & Qwen3.5-9B & \textbf{+0.22 [$-$0.14, +0.57]} & +0.11 [$-$0.20, +0.54] & +0.20 [$-$0.23, +0.50] & +0.45 [+0.23, +0.61] \\
   & Gemma-4-31B & $-$0.00 [$-$0.35, +0.46] & \textbf{$-$0.11 [$-$0.45, +0.36]} & $-$0.15 [$-$0.45, +0.38] & +0.24 [$-$0.01, +0.45] \\
   & GLM-4.6V-Flash & +0.35 [+0.09, +0.55] & +0.33 [+0.09, +0.62] & \textbf{+0.44 [+0.18, +0.65]} & $-$0.02 [$-$0.35, +0.41] \\
\bottomrule
\end{tabular}}
\end{table*}

A model might represent its weaknesses without stating them.
For each of the three models of \cref{sec:learning-weights}, we take the hidden state where its answer would begin, at the end of the prompt that asks for its accuracy on a kind (\cref{sec:before-kinds,sec:before-image}).
We take it at a quarter, half, three quarters, and the full depth of the language model and average it over the prompts of the kind.
A ridge regression maps these averages to the model's accuracy on the kinds of the training folds, with the depth and the penalty chosen by cross-validation within those folds, and predicts the kinds left out (\cref{tab:suppl-probe}).
As the second standard, we fit the same probe to the hidden states of the other two models.
On the 31 kinds of tasks, the probes find no part particular to the model, with partial correlations of \ProbeKindsQwen{} for Qwen3.5-9B, \ProbeKindsGemma{} for Gemma-4-31B, and \ProbeKindsGLM{} for GLM-4.6V-Flash, none of whose intervals lies above zero.
On image conditions, only the probe for GLM-4.6V-Flash shows this part (\ProbeCondGLM), but probes on the hidden states of the other two models show it nearly as well (\ProbeCondGLMQwenF{} and \ProbeCondGLMGemmaF).
The probe on its own hidden states exceeds them by only +0.09 [$-$0.11, +0.28], so this information is not particular to its own hidden states.
Before a model solves any question, a linear probe thus finds no more about its weaknesses in its own hidden states than in those of other models.
With about 23 kinds of tasks or 70 image conditions per fold, the probes have little data, so these results show only that we find no such information, not that it is absent.

\section{Answer probability}
\label{sec:suppl-answer}
\begin{table}[t]
\caption{The answer probability by the number of questions per axis, for the 26 open models. Part.\ is the partial correlation of \cref{sec:before-kinds} with a 95\% bootstrap interval over models, followed by the number of models above 0 and above their stated predictions. Hand-off is the score of \cref{sec:learning-decisions} when the axes are ordered by the answer probability.}
\label{tab:suppl-answer}
\centering
\scriptsize
\setlength{\tabcolsep}{2.5pt}
\resizebox{\linewidth}{!}{%
\begin{tabular}{@{}lcrrc@{}}
\toprule
Questions & Part. & $>0$ & $>$Stated & Hand-off \\
\midrule
  5 & +0.26 [+0.18, +0.34] & 23 & 17 & +0.11 [$-$0.03, +0.26] \\
  10 & +0.30 [+0.21, +0.39] & 23 & 17 & +0.15 [$-$0.02, +0.31] \\
  20 & +0.33 [+0.23, +0.43] & 22 & 17 & +0.18 [$-$0.00, +0.36] \\
  All & +0.36 [+0.25, +0.48] & 23 & 18 & +0.22 [+0.02, +0.42] \\
  \midrule
  Stated & +0.12 & & & +0.02 [$-$0.14, +0.18] \\
  Others' average & & & & +0.84 [+0.80, +0.88] \\
\bottomrule
\end{tabular}}
\end{table}

\paragraph{Questions}
The answers are those of the evaluation in \cref{sec:setup}, and we take the probability that the model assigns to the tokens of its own reply.
For five questions per axis, we draw the questions 200 times and average the partial correlation over the draws.
\Cref{tab:suppl-answer} gives the result for 5, 10, and 20 questions per axis and for all questions.
The signal grows with the number of questions but is already clear with five.
The drawn questions are also among those that define the accuracy, but this overlap does not produce the signal.
Leaving the five drawn questions out of the accuracy and the reference changes the partial correlation by less than 0.01, and taking the answer probability from a random half of the questions and the accuracy and reference from the other half gives +0.35 [+0.25, +0.45], against +0.38 when both come from the same half.

\paragraph{Reply length}
Replies differ in length between axes, so we check that the signal does not come from their length.
Removing each model's mean log reply length together with the reference changes the partial correlation from \AnswerOwn{} to \AnsLenRemoved{} [\AnsLenRemovedLo, \AnsLenRemovedHi].
The \AnsShortN{} models whose median reply has at most three tokens, which give little room for length to matter, average \AnsShortMean.

\paragraph{Resolution}
For lower resolution, we take the answer probability from 20 questions per axis and level, for the 16 models whose number of image tokens falls with the resolution, since only for them does a smaller image save cost.
The other models resize every image to the same number of tokens.

\paragraph{Decisions with the answer probability}
We also make the decisions of \cref{sec:learning-decisions} with the answer probability.
Ordering the axes by it for handing off scores 0.11 with five questions per axis and 0.22 with all questions, against 0.84 for the other models' average accuracy (\cref{tab:suppl-answer}).
Adding it to the other models' average drops changes the image budget by only \ResBudgetGain{} levels.
The answer probability thus carries information particular to the model, but it is not yet strong enough to make these decisions better than the other models' results do.

\paragraph{Putting the signal into words}
As a first attempt, we trained Qwen3.5-9B to state its accuracy on a kind after seeing five of its earlier answers on the kind, without being told whether they were right, with the folds of \cref{sec:suppl-train}.
On image conditions, its predictions then follow its particular drops more than after training on other models' accuracy (partial correlation \TOneVsCtl{} [+0.05, +0.29] higher), but Qwen2.5-VL-7B trained to read the same answers does about as well (\TOneVsReader{} [$-$0.04, +0.15]), and on kinds of tasks neither shows this signal.
For this one model, what the trained model puts into words can thus be read from its answers by another model as well, and whether a model can report more about itself than others can read from its answers remains open.

\section{Reasoning turned on}
\label{sec:suppl-think}
\begin{table}[t]
\caption{Two models with reasoning turned off and on, and the median of the open models with reasoning off. Accuracies and errors are over the 11 scored axes; the pair score and the direct choice are those of \cref{sec:before-kinds}, with the other models' average on the same pairs below the pair score; the error under image conditions is the mean difference between predicted and actual accuracy over the five axes and six levels of \cref{sec:before-image}.}
\label{tab:suppl-think}
\centering\scriptsize\setlength{\tabcolsep}{3pt}
\resizebox{\linewidth}{!}{%
\begin{tabular}{@{}lccc@{}}
\toprule
 & Qwen3.8-27B & Gemma-4-31B & Open models \\
 & off $\to$ on & off $\to$ on & (median, off) \\
\midrule
Actual accuracy & 84 $\to$ 87 & 83 $\to$ 83 & 72 \\
Stated accuracy & 91 $\to$ 85 & 92 $\to$ 89 & 87 \\
Error of stated$\downarrow$ & 12.7 $\to$ 9.3 & 10.6 $\to$ 8.2 & 18.8 \\
Pairs$\uparrow$ & 21 $\to$ 56 & 64 $\to$ 78 & 51 \\
\quad other models' average & 85 $\to$ 51 & 83 $\to$ 89 & 97 \\
Direct choice$\uparrow$ & 27 $\to$ 41 & 54 $\to$ 51 & 41 \\
Error, image conditions$\downarrow$ & 32.3 $\to$ 9.4 & 36.9 $\to$ 29.3 & 40.5 \\
\bottomrule
\end{tabular}}
\end{table}
All results in the main text are with reasoning turned off.
We reran every stage, including the evaluation of accuracy, for Qwen3.8-27B and Gemma-4-31B with reasoning turned on, in vLLM with greedy decoding and at most 16{,}384 new tokens (\cref{tab:suppl-think}).
Reasoning makes the stated numbers less overconfident and closer to the actual accuracy under image conditions, much more for Qwen3.8-27B than for Gemma-4-31B.
Reasoning does not teach the models which axis is harder for them.
Gemma-4-31B picks the harder axis in 78\% of pairs, below the other models' average (89\%).
Qwen3.8-27B moves from reversing the order of the axes (21\%) only to about chance (56\%), and still rates high-resolution search lowest although its accuracy there is above its average.
For Qwen3.8-27B the other models' average does no better (51\%), because reasoning changes its own accuracy, most on visual math (from 78\% to 90\%).
Asked directly which of two axes they would do better on, Qwen3.8-27B stays below chance (41\%) and Gemma-4-31B at chance (51\%).
The models thus still do not know which kinds of tasks are hard for them in particular when reasoning is on, although their stated numbers become less overconfident.
With only two models and no intervals, these results may not hold for other models.

\end{document}